%% file: main_v2.tex
\documentclass[journal]{IEEEtran}
\usepackage[T1]{fontenc}
\usepackage[utf8]{inputenc}
\usepackage{graphicx}
\usepackage{amsmath,amssymb}
\usepackage{array,booktabs,multirow}
\usepackage{cite}
\usepackage{url}
\usepackage{textcomp}
\usepackage{stfloats}
\usepackage{microtype}
\begin{document}

\title{SGFormer: Structure-Guided Transformer for Robust Local Feature Matching}

\author{Zhihua Xu, Runyu Zhu, and Rongjun Qin}

\maketitle

\begin{abstract}
Local feature matching is a fundamental component of photogrammetry, enabling accurate image correspondence critical for tasks such as 3D reconstruction, stereo mapping, and visual localization. While recent detector-free matching methods, like LoFTR, have advanced the field, the global features obtained by leveraging the global-range modeling capacity of the unconstrained attention mechanism compromise the model's attention to the salient structures in certain scenarios. This limitation leads to a phenomenon we define as attention divergence, wherein a portion of high-confidence matches are distributed outside the valid matching region (overlapping region), especially in scenes with large viewpoint variations. This occurs because similar features in irrelevant regions may receive equal weighting and consideration within the standard Transformer, limiting matching reliability in challenging photogrammetric environments. To address this issue in feature matching, we propose SGFormer (Structure-Guided Transformer), a novel structure-aware matching network that adaptively updates attention on features near salient structure in overlapping regions. SGFormer employs a semi-dense coarse-to-fine pipeline and incorporates the proposed Triple-Structure-Attention (TSA) module into the backbone net for extracting distinctive features. The TSA module utilizes shallow local features from early network layers to enhance the representation around salient structure, guiding subsequent transformer stages to intensify the model's focus on regions with salient structure across the global scope. SGFormer, thereby reinforcing attention to visually consistent areas while mitigating the influence of non-overlapping regions. Extensive experiments show that SGFormer significantly mitigates attention divergence and improves matching accuracy. On the MegaDepth-1500 benchmark, it achieves a mean matching accuracy of 98.2\%, outperforming existing advanced methods in both matching performance and attention concentration. On the HPatches benchmark, SGFormer achieves an MMA@0.92 and an MMA@0.98 at a 3px threshold under overall and varying illumination conditions, respectively. On the Aachen-Day-Night benchmark, SGFormer achieves a comparable localization result @95.0/92.9 at (0.5m, 5\ensuremath{{}^\circ}) threshold under day and night, respectively. These results demonstrate the robustness of SGFormer for photogrammetric tasks that demand reliable and precise image correspondences. Source code will be made publicly available.
\end{abstract}

\begin{IEEEkeywords}
Local feature matching, attention mechanism, Structure-Guided Transformer, visual localization, camera pose estimation
\end{IEEEkeywords}

\section{Introduction}

Establishing accurate and robust correspondences between images is a fundamental component in photogrammetry, essential for precise 3D reconstruction (Bullinger et al., 2021; Liu et al., 2024; Schonberger and Frahm, 2016; Widya et al., 2018; Xu et al., 2016), visual localization (Chen et al., 2024; Sattler et al., 2012), and Simultaneous Localization and Mapping (SLAM) systems (Mur-Artal and Tard\'os, 2017; Murgartal et al., 2015). Matching algorithms, therefore, require highly discriminative features and incorporate strategies that are aware of underlying scene geometry to address the inherent challenges of photogrammetric images, which often exhibit significant viewpoint changes, illumination variations, repetitive textures, and weak structural content.

Traditional detector-based methods, involving sequential keypoint detection (Luo et al., 2020; Potje et al., 2024; Revaud et al., 2019), description (Luo et al., 2018; Mishchuk et al., 2017; Tian et al., 2017; Wang et al., 2020), and matching (Cavalli et al., 2020; Sun et al., 2020; Yi et al., 2018; Zhang et al., 2019), have proven effective in structured environments (Liu et al., 2019; Sarlin et al., 2020; Tyszkiewicz et al., 2020). However, their reliance on repeatable keypoints becomes a critical bottleneck in challenging photogrammetric scenarios where such points are scarce, leading to cascading failures. Representing a paradigm shift, detector-free methods such as LoFTR (Sun et al., 2021) leverage the global receptive field of Transformer (Vaswani et al., 2017) and its variants like the Linear Transformer (Katharopoulos et al., 2020) to build dense feature representations without explicit detection. This approach has demonstrated superior performance in low-texture conditions. However, this capability can also lead to a fundamental limitation: it may impair the ability to match salient structure in certain scenarios, as the model's attention may scatter into irrelevant regions, especially under a large viewpoint change challenge---a phenomenon we refer to as attention divergence.

Fig. 1 provides a visual illustration of the attention divergence issue using the representative detector-free method, LoFTR. The correlation heatmap (Fig. 1 Column a) and the resulting putative matches (Fig.1 Column b) across different scenarios reveal a consistent problematic pattern: the model frequently scatters high-confidence points outside the true overlapping areas. For instance, in Image Pair \#1, numerous points with high-matching-confidence are erroneously distributed in non-overlapping regions (highlighted by yellow boxes in Column a), misleading the model to predict matches in invalid areas (Column b). In Image Pair \#2, a significant concentration of high-confidence points falls within regions lacking salient structures and without overlap (yellow boxes, Column a). The putative matches in Column b that may represent correct correspondences constitute only a very small proportion, significantly reducing matching accuracy. In Image Pair \#3, high-confidence points are incorrectly assigned to invalid areas such as roads and occluded regions (yellow boxes, Column a). Collectively, these results illustrate the existing issue: while the global attention mechanism in standard Transformers has been proven effective in enhancing feature matching in texture-less regions, it often scatters the model's focus into invalid matching regions (i.e., non-overlapping areas) without salient structures in certain scenarios, particularly under large viewpoint and scale variations, thereby compromising matching accuracy and reliability.

\begin{figure*}[!t]
\centering
\includegraphics[width=0.96\textwidth,trim={95pt 10pt 100pt 44pt},clip]{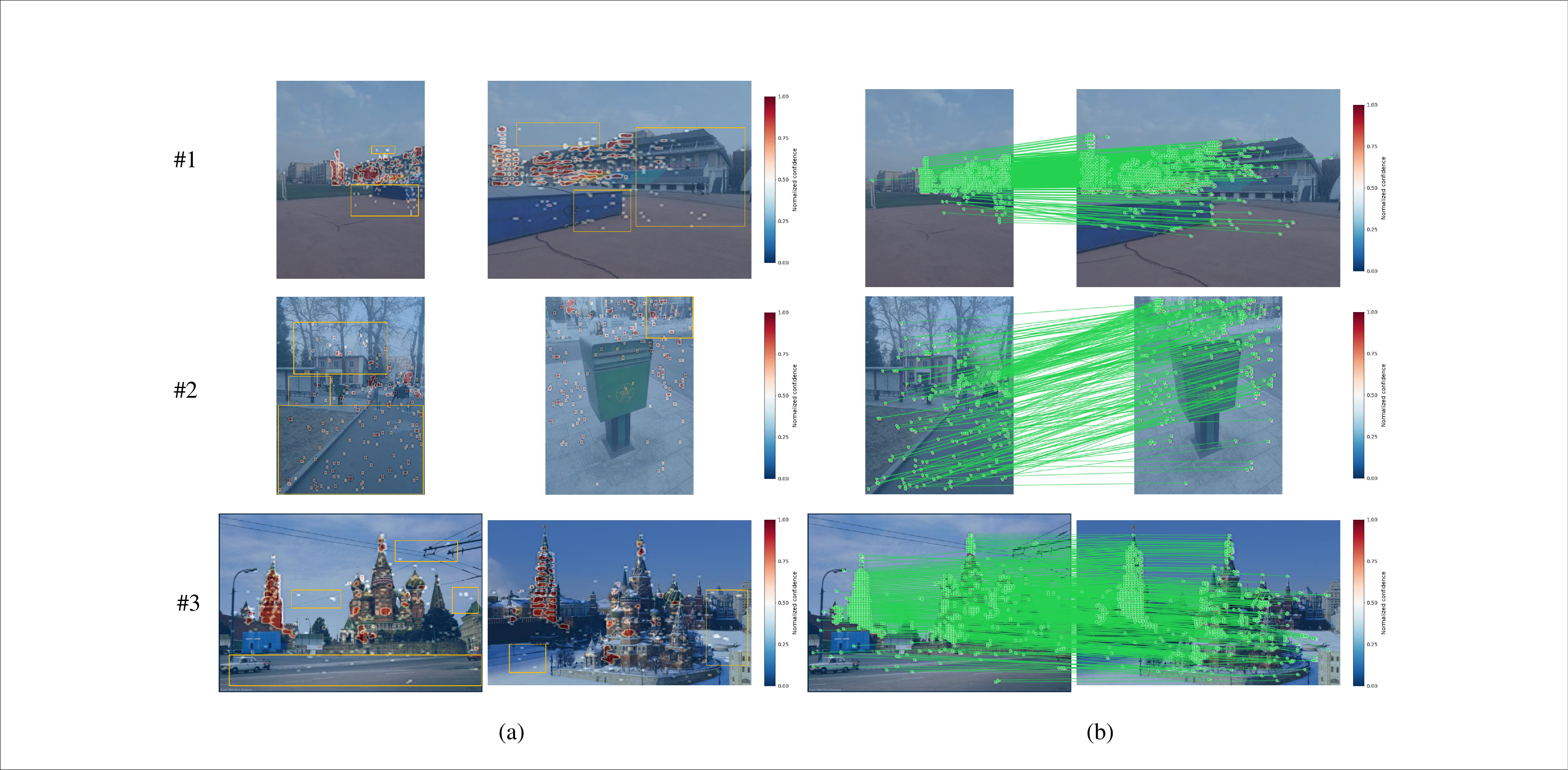}
\caption{The correlation heatmap achieved by a representative detector-free method, LoFTR (Column a), and its corresponding matches (Column b). The ``attention divergence'' phenomenon is recorded as the yellow boxes, where the scattered attention leads to mismatches in non-overlapping regions.}
\label{fig:fig1}
\end{figure*}

The above observations motivate our hypothesis: guiding the model's attention to salient structure within overlapping regions during the initial encoding stage may mitigate attention divergence and improve matching robustness. To this end, we propose SGFormer (Structure-Guided Transformer), a novel structure-aware framework for local feature matching. Its core innovation is the Triple-Structure-Attention (TSA) module, which utilizes multi-level structural cues---particularly shallow local features encoding basic image geometry---to steer the attention of subsequent deeper layers to focus on feature enhancement in the valid matching areas with salient structure. Such a structural guidance strategy effectively enhances the network's focus on features within valid regions (overlapping regions) while diminishing the influence of features from irrelevant regions. As a result, it substantially mitigates the attention divergence phenomenon induced by global attention in traditional transformers (see Appendix). Comprehensive benchmark evaluations were conducted across three datasets, demonstrating the superior capabilities of SGFormer in local feature matching. The primary contributions of this work are listed as below:

(1) The Structure-Guided Transformer (SGFormer) Framework: We propose a novel, structure-aware framework for robust feature matching. SGFormer explicitly injects spatial-structural awareness into a hierarchical Transformer architecture, leading to significant improvements in matching accuracy and robustness under challenging conditions.

(2) A novel paradigm of ''soft spatial regularization'' for global attention mechanisms: Unlike existing Transformer-based matching approaches that rely on hard region cropping or local window constraints, SGFormer incorporates low-level geometric structural priors to adaptively modulate Transformer attention. This design systematically confines the model's focus to regions of structural consistency and potential correspondences, while fully preserving the benefits of a global receptive field. The proposed paradigm establishes a new direction for mitigating attention divergence in learning-based matching framework.

(3) The Triple-Structure-Attention (TSA) Module: We design a novel attention module that adaptively guides feature aggregation within valid areas with salient structure. It combines high-resolution shallow structural features with relative spatial cues to recalibrate the attention distribution in deeper layers, ensuring focus on geometrically meaningful correspondences and effectively mitigating influence from irrelevant areas.

The remainder of this paper is organized as follows: Section 2 reviews related work. Section 3 details the architecture of SGFormer. Section 4 presents experimental results and analyses. Section 5 provides a discussion, and Section 6 concludes the study.

\section{Related work}

\subsection{Feature Point Detector and Descriptors}

Methods based on explicit feature point detection and description, commonly categorized as detector-based approaches, have long constituted the foundational paradigm for establishing local image correspondences. The classical pipeline, defined by the sequential stages of keypoint detection, local descriptor extraction, and geometric matching, was established by handcrafted features such as SIFT (Lowe, 2004), SURF (Bay et al., 2008), and ORB (Rublee et al., 2011). These methods demonstrated remarkable robustness and continue to see widespread application in various photogrammetric and 3D vision tasks. However, their performance degrades significantly in challenging scenarios, including substantial viewpoint changes, variations in illumination, and low-texture environments.

Deep learning has served as a key driver of progress in this field. Pioneering works like LIFT (Yi et al., 2016) and MagicPoint (DeTone et al., 2017) translated the principles of the traditional detector-descriptor pipeline into trainable neural networks, showing improved performance under challenging illumination and viewpoint changes. SuperPoint (DeTone et al., 2018) advanced this line by introducing a self-supervised training strategy based on homographic adaptation, yielding a more generalizable and repeatable detector-descriptor. For multimodal feature learning, ReDFeat (Deng and Ma, 2023) further recouples detection and description to improve local feature representations across heterogeneous image modalities. A landmark development, SuperGlue (Sarlin et al., 2020), redefined the matching stage by employing a graph neural network enhanced with Transformer-based attention to model contextual relationships among pre-detected keypoints, performing joint feature matching and outlier filtering in an end-to-end manner. Following this graph-based direction, DHM-Net (Chen et al., 2024) further introduces deep hypergraph modeling to capture higher-order feature relationships for robust matching. While achieving state-of-the-art results, SuperGlue and similar detector-based learning methods inherit an intrinsic limitation: their performance is fundamentally bounded by the initial detection stage. In photogrammetric scenarios characterized by extreme viewpoint changes, repetitive textures, or weak structural content, the failure to reliably detect repeatable keypoints leads to irreversible information loss and consequent matching degradation. Although modern detector-based methods (e.g., D2Net (Dusmanu et al., 2019), SuperGlue) have incorporated deeper feature representations and sophisticated matching algorithms, their dependency on a discrete set of detected points restricts their ability to leverage dense image context and limits robustness in the most challenging conditions relevant to photogrammetry.

\subsection{Detector-Free Feature Matching Methods}

To circumvent the fundamental limitations imposed by the detection stage, detector-free methods have emerged as a promising alternative. These approaches bypass explicit keypoint detection by constructing dense, pixel-wise feature representations directly from the input images. The correspondence problem is then formulated as matching within this dense feature space, enabling the utilization of all available image information and offering greater potential in texture-less or low-structure regions.

Early deep learning-based dense matching models, such as NCNet (Rocco et al., 2020b), learn correspondences end-to-end by building a 4D correlation volume and applying 4D convolutional filters to enforce neighborhood consensus. DRC-Net (Liu and Zhang, 2022) improved upon this with a coarse-to-fine optimization strategy for enhanced accuracy. A significant advancement was introduced by LoFTR, which established a widely adopted paradigm. It combines a CNN-Transformer hybrid backbone for dense feature extraction with a Transformer-based module for feature transformation and matching, followed by a coarse-to-fine matching pipeline to estimate semi-dense, sub-pixel accurate correspondences. Subsequent works have focused on refining this paradigm. ASpanFormer (H. Chen et al., 2022) integrates a differentiable optical flow estimator (Teed and Deng, 2020) to adaptively constrain the local window for cross-attention, aiming to improve efficiency and local precision. Conversely, the Quadtree Attention mechanism (Tang et al., 2022) was proposed to reduce the computational complexity of the standard Transformer (Katharopoulos et al., 2020) in LoFTR through hierarchical token pruning. For multimodal image pairs, GRiD (Liu et al., 2024) introduces guided refinement into detector-free matching, showing the value of modality-aware refinement for robust correspondence estimation. Despite these innovations, a core challenge persists: the unconstrained or globally prioritized attention mechanisms in these models can lead to attention diffusion, where the network\textquotesingle s focus is diluted across the entire image rather than concentrated on geometrically plausible overlapping regions. While ASpanFormer uses optical flow for local guidance, the reliance on deep, abstract CNN features---which often exhibit blurred spatial boundaries---can still result in misplaced matches, particularly near object contours or sky boundaries where spatial localization is critical.

A distinct direction is exemplified by MatchFormer (Wang et al., 2022), which adopts a pure Transformer architecture for unified feature extraction and matching, demonstrating competitive performance. However, this design often entails increased model complexity and computational cost. Other frameworks like COTR (Jiang et al., 2021) explore correspondence refinement in an iterative, query-driven manner, which is effective for specific tasks but less suited for dense, scene-wide matching required in photogrammetric reconstruction. Topic-assisted methods, including TopicFM (Giang et al., 2023) and TopicFM+ (Giang et al., 2024), exploit high-level semantic topics to provide contextual guidance and improve feature distinctiveness, showing that explicit context modeling can reduce ambiguity in challenging matching cases.

Alongside learning-based architectures, geometry- and optimization-driven methods remain important for robust correspondence selection. Locality-Guided Global-Preserving Optimization (Xia and Ma, 2022) removes mismatches by combining local consistency with global structure preservation, while Image Matching by Bare Homography (Bellavia, 2024) exploits local planar homography and affine constraints for reliable image matching. These methods demonstrate the effectiveness of explicit geometric regularization, but they do not directly address attention diffusion in dense detector-free Transformer matching.

Recognizing the importance of geometric constraints, another line of research explicitly estimates corresponding or overlapping regions between image pairs to guide the matching process. OETR (Y. Chen et al., 2022) exemplifies this strategy by proposing a dedicated network to regress overlap region masks as a preprocessing step. While this method demonstrates improved matching accuracy within the cropped area by eliminating gross outliers from non-overlapping zones, it introduces two significant drawbacks critical for photogrammetry: (1) it substantially increases the overall computational pipeline due to the separate preprocessing stage, and (2) the cropping operation inherently discards contextual information from areas outside the predicted mask. OAMatcher (Dai et al., 2024) integrates overlap awareness directly into the matching network. It employs a coarse matching stage to initially estimate an overlap region, which is then used to filter matches from a subsequent fine matching stage. This design improves matching precision but creates a critical dependency: the final matching quality is inextricably linked to the accuracy of the initial, often unreliable, coarse overlap prediction. Moreover, in challenging conditions with large viewpoint changes or repetitive textures, an inaccurate overlap prediction can lead to the erroneous rejection of valid matches, particularly near true boundaries, thereby undermining robustness in practical applications.

\section{Methodology}

The overall architecture of the proposed SGFormer for robust local feature matching is illustrated in Fig. 2. It follows a semi-dense, coarse-to-fine pipeline and consists of three core components: (1) Hierarchical Structure-Guided Transformer Backbone for overlapping enhanced multi-scale feature extraction, (2) Coarse-level Matching Module for initial correspondence estimation, and (3) Fine-level Matching Module for sub-pixel refinement. The process begins by feeding an image pair into the backbone network. This backbone, detailed in Section 3.1, is a four-stage hierarchical Transformer explicitly designed to inject spatial-structural guidance. It outputs two levels of feature representations: lower-resolution coarse-level and higher-resolution fine-level feature maps. The coarse-level features from both images are then passed to the matching module (Section 3.2), which establishes a set of coarse semi-dense matches. Finally, these coarse matches are precisely localized to sub-pixel accuracy by the fine-level module (Section 3.3), leveraging the finer-grained feature maps. The coarse and fine matching modules maintain a design consistent with the established LoFTR (Sun et al., 2021) pipeline for fair comparison, allowing the analysis to focus on the contributions of the novel backbone. The following sections elaborate on the design and implementation of each component.

\begin{figure*}[!t]
\centering
\includegraphics[width=0.96\textwidth]{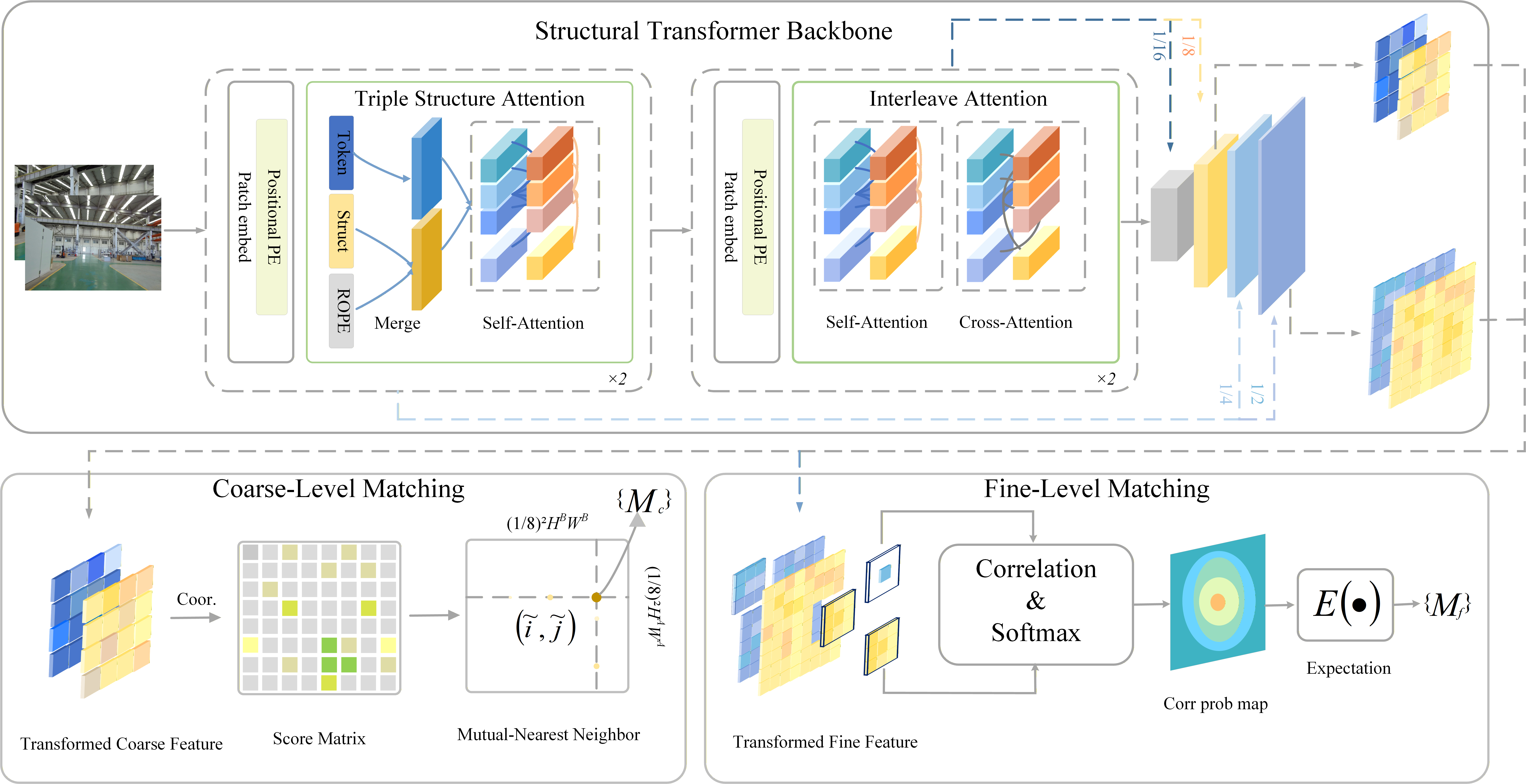}
\caption{Framework of the proposed SGFormer for local feature matching.}
\label{fig:fig2}
\end{figure*}

\subsection{Hierarchical Structural-Guided Transformer Backbone}

The core of SGFormer is a hierarchical, four-stage Transformer backbone designed to explicitly integrate structural guidance for robust feature representation. As illustrated in Fig. 2, the first two stages incorporate the novel Triple-Structure-Attention (TSA) module to enforce an initial focus on geometrically salient overlapping regions. The latter two stages employ standard Interleaved Self-/Cross-Attention blocks to further refine features with global context. Finally, a Feature Pyramid Network (FPN)-style decoder fuses multi-scale features to output dense and discriminative representations at both coarse (1/8 original resolution) and fine (1/2 original resolution) levels for subsequent matching.

\subsubsection{Positional Patch Embedding}

Preserving accurate spatial relationships is critical for feature matching. Standard transformer tokenization discards inherent image structure, and while additive positional encodings (PEs) can restore some spatial awareness, they risk introducing a strong positional bias. In regions with weak textures, this bias can cause the model to overfit to positional coincidences rather than genuine visual correspondence, ultimately degrading generalization.

To mitigate this limitation, we adopt the Positional Patch Embedding (PosPE) module from MatchFormer (Wang et al., 2022) as a front-end component to the transformer block. The process begins with a 7$\times$7 convolutional layer (stride 2, padding 3) in the first stage, followed by 3$\times$3 convolutions (stride 2, padding 1) in subsequent stages. A key component is a depth-wise separable convolution that extracts low-level features to generate pixel-wise, content-aware weights via a sigmoid function \(\sigma( \bullet )\). These weights modulate the standard positional encoding, allowing the model to adaptively balance spatial and visual cues. This design suppresses positional overfitting in ambiguous regions and enhances the geometric integrity of the resulting feature tokens.

\subsubsection{Triple-Structural-Attention Module}

The TSA module is the pivotal innovation designed to counteract the attention divergence problem identified in Fig. 1. It is motivated by three principles: (1) Geometric Structure Guidance: Human perception relies on elementary structures (edges, corners); similarly, providing explicit geometric cues can steer the network's attention. (2) Relative Spatial Context: Precise matching requires understanding relative positions between features; thus, explicitly encoding relative positions along the channel dimension into structural features enhances spatial context modeling. (3) Utilization of Shallow Features: Shallow convolutional layers capture high-resolution structural details ideal for providing guidance, despite having limited semantic abstraction.

The TSA module, depicted in Fig. 3, is a novel design that helps the network focus on corresponding regions with similar structures. It operates as follows. First, a light-weight, non-pretrained structural feature extractor is applied to the input feature map. This extractor uses depth-wise separable convolutions and a sigmoid activation to produce a normalized structural intensity map \(f_{struct}\), rich in edge and texture information, without adding high computational cost:

\begin{equation}
\begin{aligned}
f_{struct} = \operatorname{sigmoid}\bigl(&\operatorname{Conv}_{3\times3}( \\
&\operatorname{LeakyReLU}(\operatorname{BN}(\operatorname{DSConv}(f))))\bigr)
\end{aligned}
\tag{1}\label{eq:eq1}
\end{equation}

\begin{equation}
DSConv = {Conv}_{1*1}({depthwiseConv}_{3*3}( \bullet ))\tag{2}\label{eq:eq2}
\end{equation}

\noindent where \({Conv}_{n*n}\) means convolution with \(n*n\) kernel. \(DSConv\) stands for a depth-wise separable convolution. BN stands for batch normalization.

To integrate spatial information without the bias of additive encodings and let the model adaptively balance the weight between spatial info and structural feature, Rotational Relative Positional Encoding (RoPE) is incorporated to model spatial relativity. The structural feature \(f_{struct}\) and RoPE are explicitly concatenated along the channel dimension and projected to form a structure token \(T_{s}\):

\begin{equation}
T_{s} = flatten({Conv}_{1*1}(concat(f_{struct},\ ROPE(f))))\tag{3}\label{eq:eq3}
\end{equation}

Simultaneously, the input visual feature is projected to a vision token \(T_{v}\). Within the attention block, \(T_{s}\) provides the Query and Key, while \(T_{v}\) provides the Value. This design ensures that attention is computed based on structural similarity, effectively re-weighting the visual features:

\begin{equation}
\begin{aligned}
T_{v}^{w} = \operatorname{LinearAttention}_{multi}\bigl(&Q = T_{s},\ K = T_{s}, \\
&V = T_{v}\bigr)
\end{aligned}
\tag{4}\label{eq:eq4}
\end{equation}

The weighted vision token \(T_{v}^{w}\) is then concatenated with the original \(T_v\) and fused via a depth-wise convolutional multi-layer perceptron (MLP), followed by a residual connection to produce the final output:

\begin{equation}
T_{out} = T_{v} + {MLP}_{dwconv}(concat(T_{v},\ T_{v}^{w}))\tag{5}\label{eq:eq5}
\end{equation}

\noindent where \(T_{v}\) and \(T_{s}\) represent \({Token}_{vision}\) and \({Token}_{struct}\), respectively. The \({MLP}_{dwconv}\) can be formulated as Eq. (6):

\begin{equation}
{MLP}_{dwconv} = fc(LeakyRelu({depthwiseConv}_{3*3}(fc( \cdot ))))\tag{6}\label{eq:eq6}
\end{equation}

\noindent where \(fc\) denotes a fully connected layer. TSA directs attention toward structurally similar regions, enhancing correspondence-relevant representations. Critically, RoPE is exclusively applied in TSA modules. Compared to architectures like MatchFormer (three attention layers per block), TSA uses a single self-attention layer, reducing computational overhead.

\begin{figure}[!t]
\centering
\includegraphics[width=0.88\columnwidth]{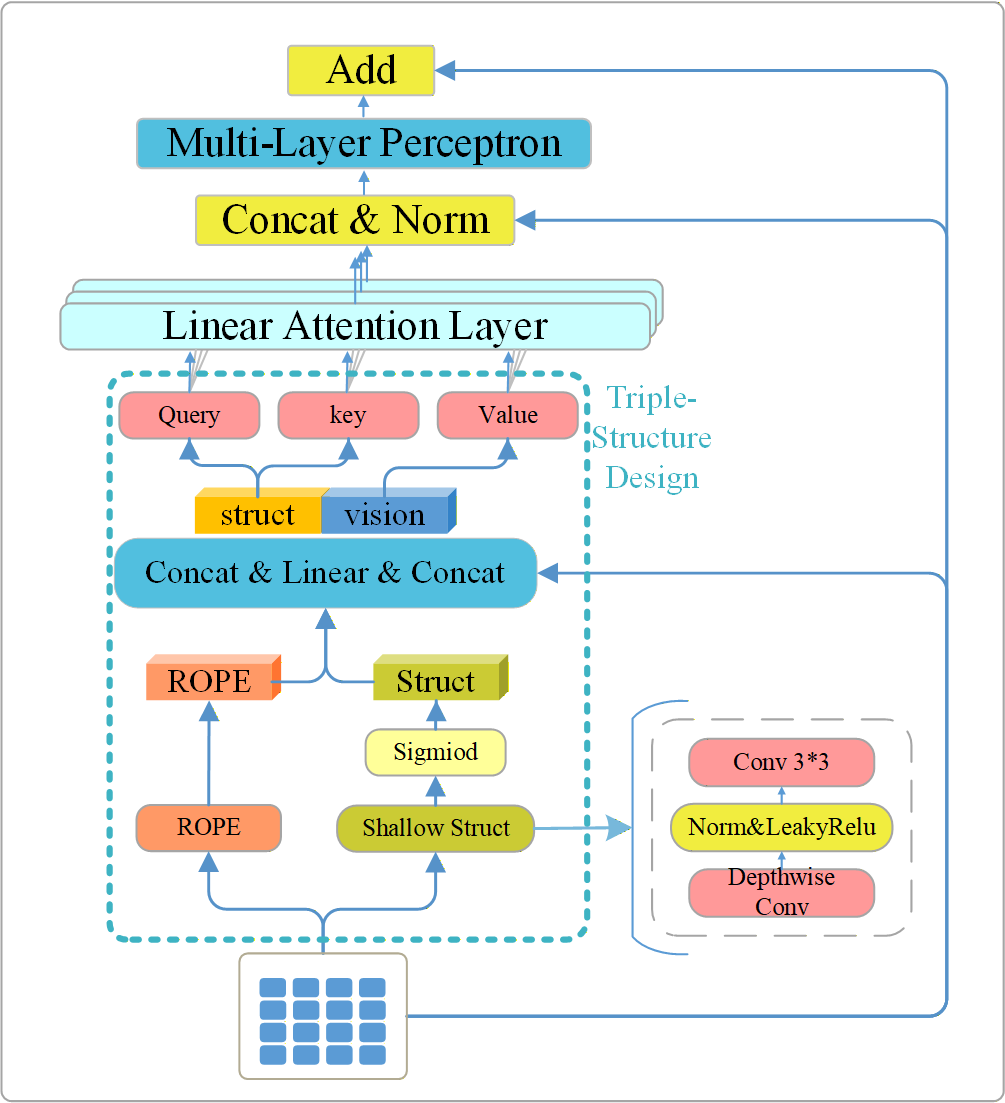}
\caption{Triple Structure Attention Module.}
\label{fig:fig3}
\end{figure}

\subsubsection{Interleaving Self-/Cross- Attention}

Inspired by LoFTR (Sun et al., 2021), the last two stages of the backbone employ interleaved blocks of standard (Vanilla) self-attention and cross-attention. While linear attention is efficient, vanilla attention offers superior representational capacity needed for high-level semantic refinement. These stages operate on lower-resolution features (1/8 and 1/16) that have already been primed by the TSA modules. The self-attention layers enhance intra-image context, while the cross-attention layers model inter-image dependencies. This progression mimics a logical matching strategy: initial coarse localization via structural guidance (TSA stages), followed by semantic verification and refinement through global contextual modeling.

\subsubsection{Multi-Scale Feature Fusion}

To leverage information from all stages, an FPN-style decoder aggregates the multi-scale feature maps. This decoder performs iterative upsample and fusion, combining high-resolution, structurally rich features from early stages with semantically rich and discriminative features from deeper stages. The output is a pair of refined, multi-scale feature pyramids that provide both the broad context necessary for robust coarse matching and the fine details required for precise sub-pixel localization in the subsequent modules.

\subsection{Coarse-level Matching Module}

The coarse-level matching module aims to establish an initial set of semi-dense correspondences between images using the transformed coarse feature maps \(F_{A}^{c}\) and \(F_{B}^{c}\) output by the Structure-Guided Transformer backbone.

\begin{figure}[!t]
\centering
\includegraphics[width=0.98\columnwidth]{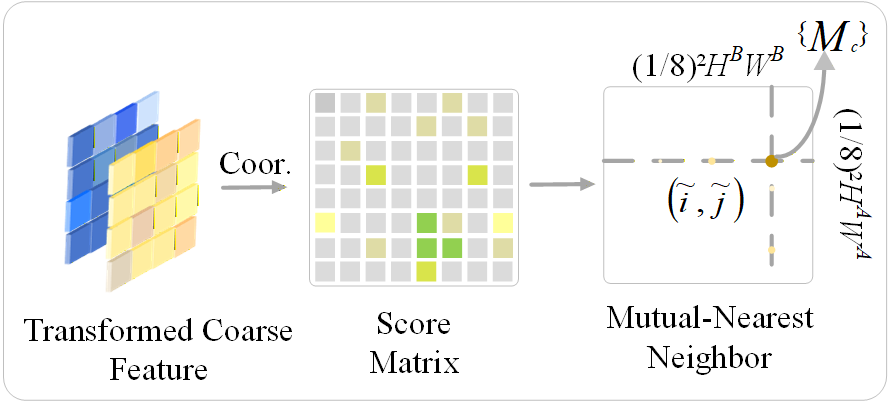}
\caption{Coarse-Level Matching module pipeline}
\label{fig:fig4}
\end{figure}

\begin{equation}
P_{c} = {softmax}_{row}(S) \cdot {softmax}_{col}(S)\tag{7}\label{eq:eq7}
\end{equation}

\begin{equation}
S = \sigma F_{A}^{c}{F_{B}^{c}}^{T} \in R^{n \times m}\tag{8}\label{eq:eq8}
\end{equation}

The coarse correspondences, denoted as \(\mathcal{M}_{c}\), are further filtered using a correlation threshold \(\tau\) and subjected to the mutual-nearest-neighbor (MNN) constraint. The prediction process is formulated as Eq. (9).

\begin{equation}
\mathcal{M}_{c} = \{(i^{c},j^{c})|\forall\left( i^{c},j^{c} \right) \in MMN\left( P_{c} \right),\ P_{c}(i^{c},j^{c}) \geq \tau\}\tag{9}\label{eq:eq9}
\end{equation}

\subsection{Fine-level Matching Module}

Building upon the coarse correspondences, the fine-level matching module refines their locations to higher-resolution (1/2 original resolution) feature maps from the backbone and calculates an expectation over the distribution of matching probability to get a high-level accuracy. This process, illustrated in Fig. 5, enhances geometric precision critical for photogrammetric applications such as dense 3D reconstruction.

\begin{figure}[!t]
\centering
\includegraphics[width=0.98\columnwidth]{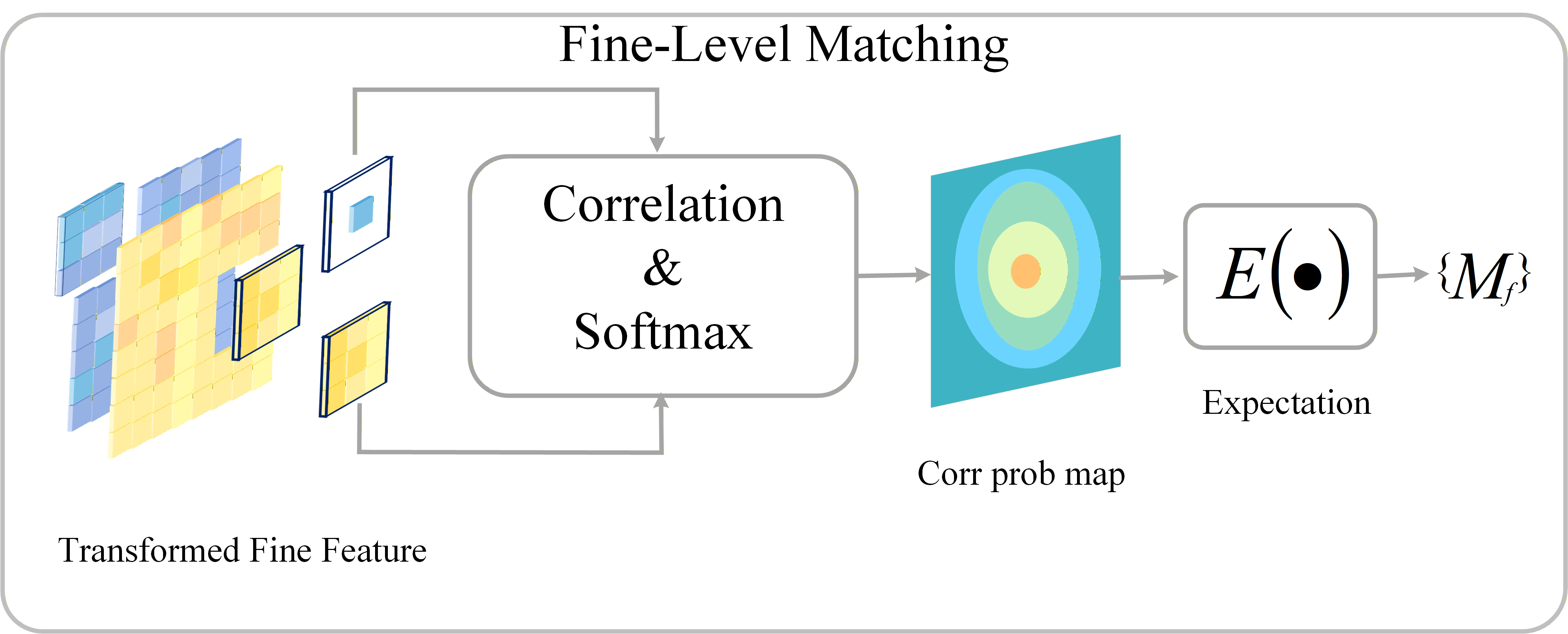}
\caption{Fine-Level Matching module pipeline}
\label{fig:fig5}
\end{figure}

For each coarse correspondence \((i^{c},j^{c})\) in the set \(\mathcal{M}_{c}\), corresponding local patches (spans), i.e., \(F_{span}^{A}\left( i^{f} \right),\ F_{span}^{B}\left( j^{f} \right)\) of size \(w \times w\), are cropped from the fine feature maps of both images. Unlike (H. Chen et al., 2022; Sun et al., 2021), no additional transformer blocks are employed for further feature aggregating, as this incurs significant computational overhead with marginal performance gains. Finally, the center vector of each patch \(F_{span}^{A}\left( i^{f} \right)\) in the source image is correlated with all feature vectors \(F_{span}^{B}\left( j^{f} \right)\) within the corresponding target patch. A matching probability map is computed from this correlation, and the final sub-pixel correspondence \((i^{f},j^{f^{'}})\) is determined by calculating the expectation over this probability distribution. Aggregating all refined matches \(\{(i^{f},j^{f^{'}})\}\) yields the final set of fine-level matches, \(\mathcal{M}_{f}\).

This efficient yet effective refinement strategy aggregates all processed pairs \{\((i^{f},j^{f^{'}})\)\} to produce the final set of high-precision correspondences \(\mathcal{M}_{f}\). The module capitalizes on the enhanced localization properties of SGFormer's fine-level features, translating robust coarse matches into accurate sub-pixel measurements suitable for demanding photogrammetric pipelines.

\subsection{Supervision}

The entire SGFormer pipeline is trained end-to-end under a composite loss function that jointly supervises the coarse and fine matching stages. This ensures the network learns to produce both reliable initial correspondences and high-precision refinements. The total loss \(\mathcal{L}\) is defined as:

\begin{equation}
\mathcal{L = \ }\mathcal{L}_{c} + \mathcal{L}_{f}\tag{10}\label{eq:eq10}
\end{equation}

\noindent where \(\mathcal{L}_{c}\) and \(\mathcal{L}_{f}\) are the coarse-level and fine-level losses, respectively.

\subsubsection{Coarse-Level Supervision}

Following Sun et al. (2021), ground-truth for coarse-level supervision is generated photogrammetrically. Using the available camera poses and depth maps, pixel coordinates from image \(I^{A}\) are projected onto image \(I^{B}\) at the coarse feature resolution (1/8 original resolution) to obtain the set of ground-truth correspondences \({\{\mathcal{M}_{c}\}}_{gt}\). Supervision is enforced by minimizing the negative log-likelihood loss over the predicted probability matrix \(\mathbf{P}_{c}\),

\begin{equation}
\mathcal{L}_{c} = - \frac{1}{|{\{\mathcal{M}_{c}\}}_{gt}|}\sum_{(i^{c},j^{c})\epsilon{\{\mathcal{M}_{c}\}}_{gt}}^{}{\log{P_{c}(i^{c},j^{c})}}\tag{11}\label{eq:eq11}
\end{equation}

This objective encourages the model to assign high matching confidence to geometrically valid correspondence.

\subsubsection{Fine-Level Supervision}

For fine-level refinement, supervision aims to minimize the sub-pixel localization error. The ground-truth fine correspondence \({j^{f^{'}}}_{gt}\) for a predicted point \(j^{f}\) is derived analogously to the coarse level, using the same camera geometry.

Adopting the strategy from Sun et al. (2021), we treat the fine-level loss \(\mathcal{L}_{f}\) as a variance-weighted \(\mathcal{l}_{2}\) loss, which adaptively penalizes predictions based on the uncertainty of the estimated probability distribution:

\begin{equation}
\mathcal{L}_{f} = \frac{1}{|\mathcal{M}_{f}|}\sum_{(i^{f},j^{f^{'}}) \in \mathcal{M}_{f}}^{}{\frac{1}{\sigma^{2}\left( i^{f} \right)}\left\| j^{f^{'}} - {j^{f^{'}}}_{gt} \right\|_{2}}\tag{12}\label{eq:eq12}
\end{equation}

\noindent where \(\sigma^{2}(i^{f})\) denotes the total variance of the predicted probability distribution for point \(i^{f}\). This weighting scheme reduces the influence of high-uncertainty matches during training, leading to more stable optimization and a model that is more confident in its accurate predictions.

\section{Experimental Results}

\begin{table*}[!t]
\caption{Datasets used for evaluation.}
\label{tab:table1}
\centering
\input{tables/table01.tex}
\end{table*}

To comprehensively evaluate the performance of SGFormer, we conducted experiments on three publicly available benchmarks widely used in image matching and photogrammetry: the HPatches dataset (Balntas et al., 2017) for assessing general matching accuracy, the MegaDepth dataset (Li and Snavely, 2018) for evaluating relative pose estimation, and the Aachen-Day-Night dataset (Sattler et al., 2018) for testing visual localization. The key characteristics and purposes of these datasets are summarized in Table~\ref{tab:table1}.

\subsection{Implementation Details}

The SGFormer model was trained on the MegaDepth dataset (Li and Snavely, 2018). Following the data partition protocol of LoFTR (Sun et al., 2021), we used the official training and validation splits. Due to missing depth information in Scene 148 of the downloaded MegaDepth dataset, we excluded it, resulting in a final training set of 38,000 raw images from 365 scenes (without D2-Net pre-processing). All images were resized to a resolution of 832 $\times$ 832 pixels. We employed the AdamW optimizer with an initial learning rate of \(4 \times 10^{- 4}\), a linear warm-up over the first 3 epochs, and a batch size of 1 for each GPU. The entire training process took approximately 40 hours on four NVIDIA GeForce RTX 4090D GPUs. For inference, a single NVIDIA RTX A6000 (48GB) GPU was used.

\subsection{Evaluation and Comparison}

\subsubsection{Evaluation Metrics}

In the local feature matching task, we used Mean Matching Accuracy (MMA) and Area Under the Curve (AUC) to evaluate the matching performance of matching methods (H. Chen et al., 2022; Sun et al., 2021). MMA measures the average percentage of correct matches across a range of pixel error thresholds, which quantifies the geometric correctness and robustness of feature matches. AUC measures the area under the ``accuracy vs. error threshold'' curve. It evaluates an algorithm's performance comprehensively across all possible error thresholds.

Besides, with the objective of evaluating the attention divergence phenomenon through matching confidence heatmaps, we introduced three indicators to assess matching performance on the Aachen-Day-Night and MegaDepth datasets, where image pairs exhibit large viewpoint variances. We fit an ellipse based on the matching confidence for each testing image pair to describe the spatial distribution of matches. We calculated a weighted centroid of the ellipse using the 2D locations of high-confidence points (confidence \textgreater{} 0.80). Then, translating the whole points (confidence is not zero) into the coordinate system with the weighted centroid as the origin. With that, we projected the centralized points to two principal component directions, and calculate the standard deviation in the direction of the two principal components represented by \(std_{x}\) and \(std_{y}\). Based on the ellipse, we calculated ellipse\_area (\(ea\)), coverage\_1sigma (\(1Sig\)) and mean\_distance\_to\_center (\(md2c\)) to comprehensively measure the attention divergence. The formulation of these three indicators is defined as:

\begin{equation}
ea = \pi*std_{x}*std_{y}\tag{13}\label{eq:eq13}
\end{equation}

\begin{equation}
1Sig = \frac{1}{N}\sum_{i = 1}^{N}\left\{ \begin{array}{r}
1,(\frac{x_{i}^{'2}}{{std_{x}}^{2}} + \frac{y_{i}^{'2}}{{std_{y}}^{2}} \leq 1) \\
0,(\frac{x_{i}^{'2}}{{std_{x}}^{2}} + \frac{y_{i}^{'2}}{{std_{y}}^{2}} > 1)
\end{array} \right.\ \tag{14}\label{eq:eq14}
\end{equation}

\noindent where \(x_{i}^{'}\), \(y_{i}^{'}\) are formulated as:

\begin{equation}
x_{i}^{'} = \left( x_{i} - c_{x} \right)\cos\theta + (y_{i} - c_{y})\sin\theta\tag{15}\label{eq:eq15}
\end{equation}

\begin{equation}
y_{i}^{'} = - \left( x_{i} - c_{x} \right)\sin\theta + (y_{i} - c_{y})\cos\theta\tag{16}\label{eq:eq16}
\end{equation}

\noindent where \((c_{x},c_{y})\) is the center of the ellipse, and \(\theta\) is getting from PCA.

\begin{equation}
md2c = \frac{1}{N}\sum_{i = 1}^{N}\sqrt{{(x_{i} - c_{x})}^{2} + {(y_{i} - c_{y})}^{2}}\tag{17}\label{eq:eq17}
\end{equation}

For ellipse\_area (\(ea\)) and mean\_distance\_to\_center (\(md2c\)), lower values mean a more aggregated high confidence area and a less dispersed distribution of matching confidence, respectively. For coverage\_1sigma (\(1Sig\)), higher values indicate that more points are distributed in the high confidence area, which may mean a better distribution under a large viewpoint change scene. All of these three indicators should be considered as a whole.

Besides, we introduced High Confidence Ratio (\(HCR\)):

\begin{equation}
HCR = \frac{1}{N}\sum_{i - 1}^{N}\left\{ \begin{array}{r}
1,\ (c_{i} > T) \\
0,\ (c_{i} < T)
\end{array} \right.\ \tag{18}\label{eq:eq18}
\end{equation}

\noindent where \(N\) is the total number of pixels in an image, \(c_{i}\) is the confidence of pixel \(i\), and \(T\) is the confidence threshold. Higher \(HCR\) indicates the wider coverage of high matching confidence.

\begin{table*}[!t]
\caption{Mean Matching Accuracy result on the HPatches dataset.}
\label{tab:table2}
\centering
\input{tables/table02.tex}
\vspace{1pt}
\parbox{\linewidth}{\footnotesize\emph{Note:} The highest scores in each column are highlighted in bold.}
\end{table*}

\subsubsection{Image Matching on HPatches}

We evaluated matching robustness on the HPatches dataset, which contains 52 sequences with illumination changes and 56 sequences with viewpoint changes. The proposed SGFormer was compared against four detector-based methods (D2-Net (Dusmanu et al., 2019), ASLFeat (Luo et al., 2020), CAPS\_SIFT+NN, SuperPoint (DeTone et al., 2018) + SuperGlue (Sarlin et al., 2020)) and five detector-free, semi-dense methods (COTR(Jiang et al., 2021), Sparse-NCNet (Rocco et al., 2020a), LoFTR (Sun et al., 2021), Patch2Pix (Zhou et al., 2021). Following the standard protocol as LoFTR (Sun et al., 2021), input images were resized to 480 pixels on the shortest side. Matching performance was quantified by Mean Matching Accuracy (MMA) at 1px, 3px, and 5px thresholds, respectively.

\begin{figure*}[!t]
\centering
\includegraphics[width=0.96\textwidth]{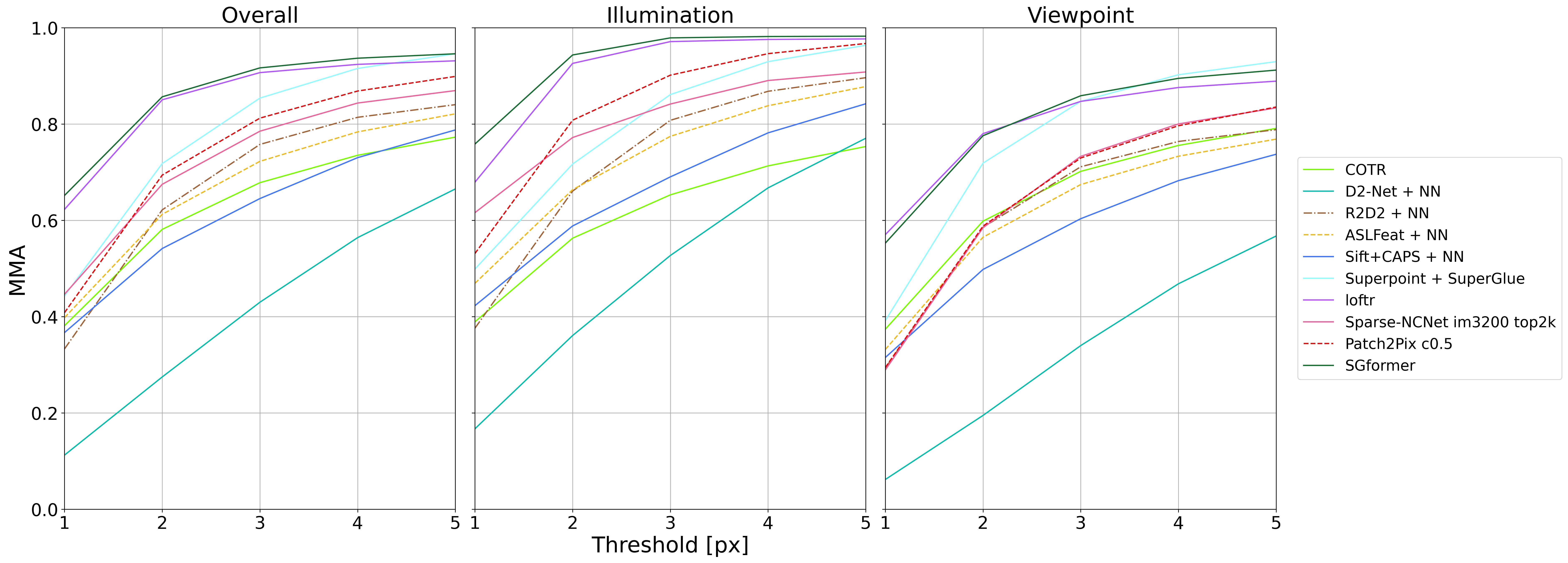}
\caption{Image matching on the HPatches dataset.}
\label{fig:fig6}
\end{figure*}

The quantitative results are summarized in Table~\ref{tab:table2} and visualized in Fig. 6. SGFormer achieves the best overall MMA scores across all thresholds. Notably, it outperforms all detector-based methods, especially at stricter thresholds, highlighting the advantage of the detector-free paradigm. Among detector-free methods, SGFormer attains the best performance, followed by LoFTR. Under overall conditions in the HPatches dataset, SGFormer surpasses LoFTR by 4.84\%, 1.10\%, and 2.15\% at the 1px, 3px, and 5px thresholds, respectively.

A detailed breakdown reveals that under illumination changes, SGFormer's advantage is even more pronounced (e.g., +11.76\% at 1px, and 2.08\% at 3px). This demonstrates the exceptional robustness of its structure-guided attention against photometric distortions, as the TSA module prioritizes geometric consistency over appearance.

Under viewpoint changes, SGFormer maintains strong performance but does not outperform all methods at every threshold (e.g., slightly lower than LoFTR at 1px, and SuperPoint+SuperGlue at 5px). This indicates a limitation: severe geometric distortions and non-rigid deformations can reduce the reliability of the structural consistency prior used for guidance. Nevertheless, SGFormer still achieves the best performance at the 3px threshold, confirming its overall efficacy.

Fig. 7 visualizes the correlation heatmaps for LoFTR and SGFormer on three representative scenes from HPatches dataset. Table~\ref{tab:table3} shows the quantization results for \(HCR\), corresponding to the scenes in Fig. 7. The comparison provides direct evidence of how SGFormer mitigates the attention divergence problem.

In scenes with only illumination change, higher \(HCR\) with fewer points distributed at invalid area should be considered better. In \#1 of Fig. 7 (severe illumination change), LoFTR\textquotesingle s matching confidence is sparse and weak, and the matching confidence only cover 20.4\% (\(HCR\)). In contrast, SGFormer, guided by structural similarity, produces a denser (\(HCR\)@55.4\%) and more confident heatmap concentrated on the overlapping structure, effectively overcoming the photometric challenge. In \#2 \& \#3, LoFTR reaches a better coverage at \(HCR\)@59.3\% \& @52.4\% than in \#1, but it exhibits clear attention divergence, with spurious high-confidence responses in non-overlapping areas (e.g., the sky), which would lead to outliers. SGFormer\textquotesingle s heatmaps are more concentrated within the true available regions with \(HCR\)@64.5\% \& @54.1\%, demonstrating superior focus and a reduction in the dispersion of high confidence matches.

\begin{figure*}[!t]
\centering
\includegraphics[width=0.96\textwidth,trim={83pt 20pt 90pt 32pt},clip]{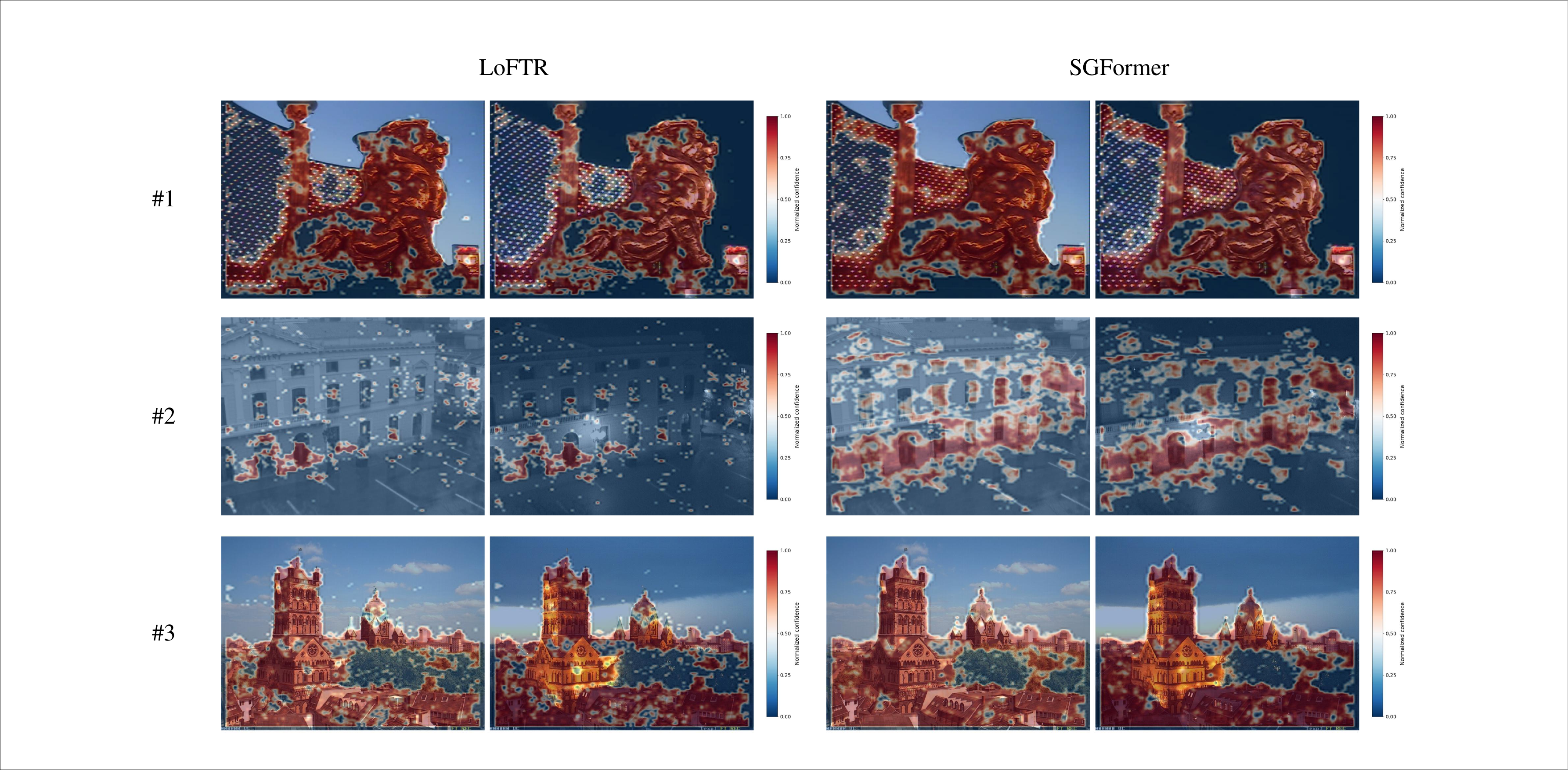}
\caption{Correlation heatmaps of LoFTR (left) and SGFormer (right). Warmer color indicates higher matching confidence.}
\label{fig:fig7}
\end{figure*}

\begin{table}[!ht]
\caption{comparison on the High Confidence Ratio (HCR) between LoFTR and SGFormer corresponds to scenes in Fig. 7.}
\label{tab:table3}
\centering
\input{tables/table03.tex}
\end{table}

These visual results in Fig. 7 and Table~\ref{tab:table3} confirm that the proposed SGFormer concentrates more attention toward geometrically plausible regions, leading to more reliable matching confidence, which is reflected in the superior accuracy metrics of Table~\ref{tab:table2}.

\subsubsection{Visual Localization on Aachen-Day-Night}

Visual localization, which estimates the 6-degree-of-freedom (6-DoF) camera poses of a query image within a pre-built 3D model, is a fundamental photogrammetric task for mapping and navigation. Its accuracy is critically dependent on the robustness of feature matching, especially under challenging conditions such as large viewpoint changes and day-night illumination variations. We evaluated this capability using the Aachen-Day-Night benchmark, integrating SGFormer into the standard HLoc pipeline (Sarlin et al., 2019). Comparisons are made against two detector-based methods (SuperPoint (DeTone et al., 2018) + SuperGlue (Sarlin et al., 2020) and D2-Net (Dusmanu et al., 2019)), and four detector-free semi-dense approaches (LoFTR (Sun et al., 2021), COTR (Jiang et al., 2021), ASpanFormer (H. Chen et al., 2022), and TopicFM (Giang et al., 2023)). Following Long-Term Visual Localization Benchmark standards, success rates were measured by the area under the cumulative curve (AUC) of pose errors within increasingly relaxed thresholds (0.25m, 2$^\circ$), (0.5m, 5$^\circ$), and (1m, 10$^\circ$), respectively.

\begin{table*}[!t]
\caption{Visual localization results on the Aachen-Day-Night dataset.}
\label{tab:table4}
\centering
\input{tables/table04.tex}
\vspace{1pt}
\parbox{\linewidth}{\footnotesize\emph{Note:} Best results are in bold; second best are underlined. Some results for D2-Net at strict thresholds are not publicly reported (``-'').}
\end{table*}

Table~\ref{tab:table4} shows the quantitative results on the Aachen-Day-Night benchmark. SGFormer demonstrates highly competitive and robust performance. During the day, it achieves the second-highest accuracy at the medium precision threshold (AUC@(0.5m, 5$^\circ$)), outperforming ASpanFormer by 0.9 percentage points. At night, SGFormer matches the best performance at the most stringent threshold (AUC@(0.25m, 2$^\circ$)), which is crucial for high-precision applications. While its performance at the loosest thresholds (1m, 10$^\circ$) is slightly lower than the best methods, SGFormer consistently delivers superior or on-par accuracy in the high-precision regime that is most relevant for advanced visual localization.

\begin{figure*}[!t]
\centering
\includegraphics[width=0.96\textwidth,trim={72pt 27pt 120pt 22pt},clip]{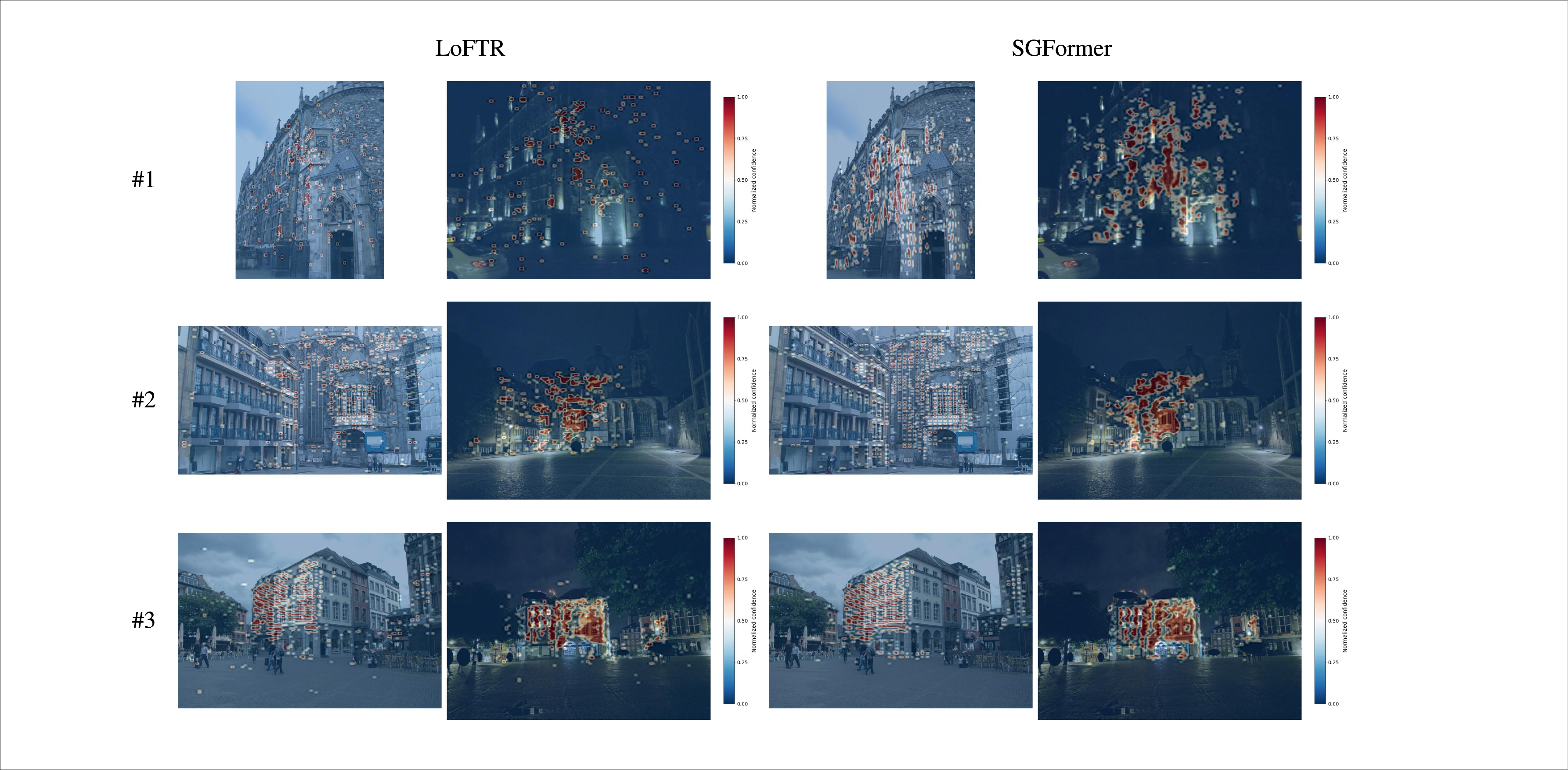}
\caption{correlation heatmaps of LoFTR (left) and SGFormer (right) on Aachen images. Warmer color indicates higher matching confidence.}
\label{fig:fig8}
\end{figure*}

\begin{table}[!t]
\caption{The quantitative result of scenes corresponds to Fig. 8}
\label{tab:table5}
\centering
\input{tables/table05.tex}
\end{table}

Figs. 8 and 9 present a qualitative comparison of the matching behaviors between LoFTR and SGFormer on sample Aachen image pairs.~This visual analysis is supplemented by the quantitative evaluation presented in Table~\ref{tab:table5}, which reports the \(HCR,ea,1Sig\), and \(md2c\) indicators. Due to the distinctive features of the architectural area, both methods demonstrated relatively more concentrated attention heatmap distributions in regions with distinct geometric structures. However, LoFTR exhibited a noticeable attention divergence phenomenon. The correlation heatmaps in Fig. 8 and the quantitative result in Table~\ref{tab:table5} reveal such a key difference: LoFTR frequently exhibits attention divergence, with spurious high-confidence responses scattered in non-overlapping regions (e.g., sky, distant buildings). These false positives, evident in \#1 to \#3, mislead the matching process. In contrast, SGFormer's heatmaps are more concentrated and comprehensive within the true overlapping areas.

\begin{figure*}[!t]
\centering
\includegraphics[width=0.96\textwidth,trim={110pt 20pt 82pt 22pt},clip]{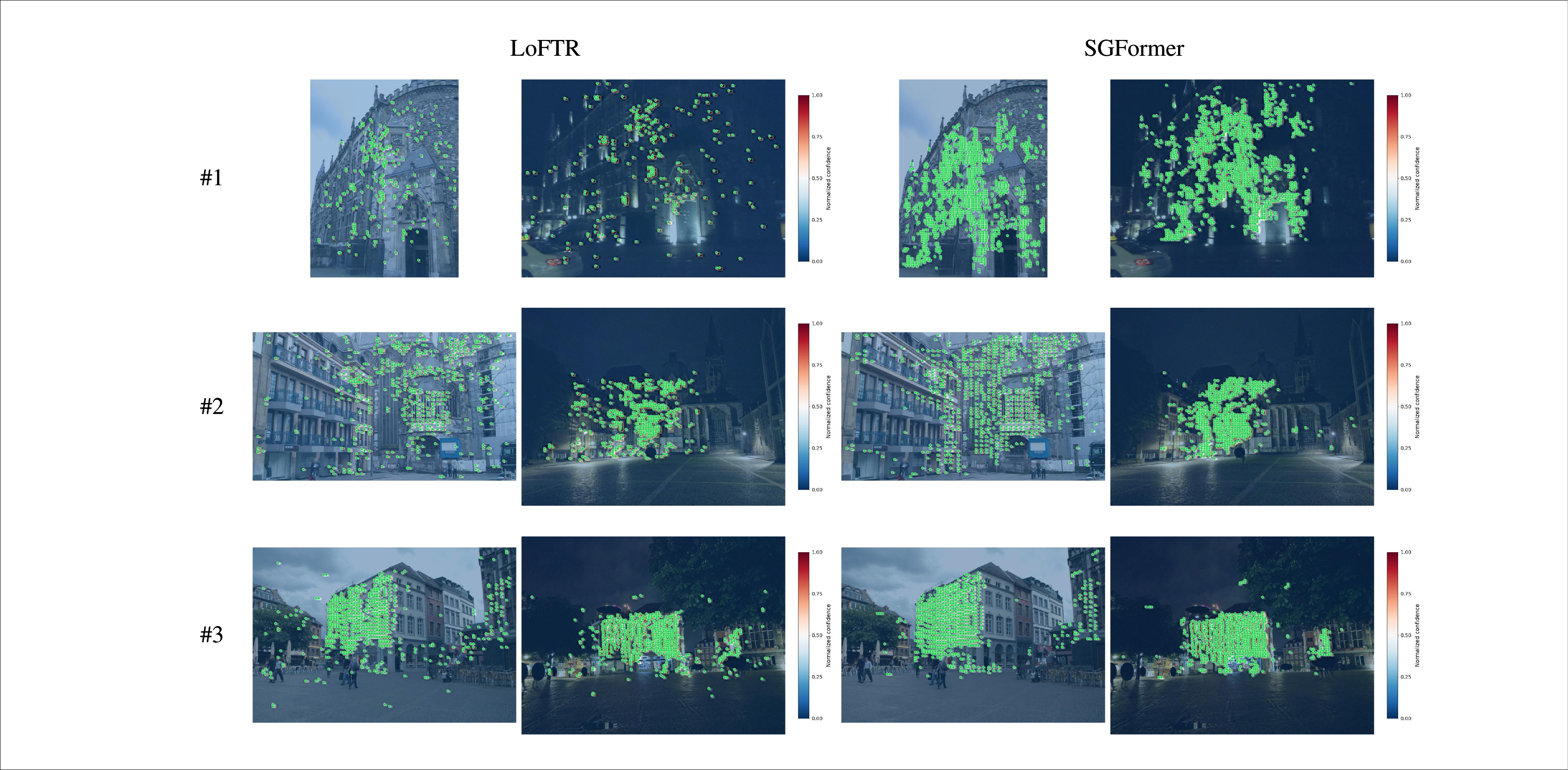}
\caption{visualization of matches corresponding to correlation heatmaps in Fig. 8.}
\label{fig:fig9}
\end{figure*}

In \#2 and \#3, the area of the fitted ellipse (\(ea\)) for SGFormer decreased by 13.4\% and 16.8\%, respectively, compared to LoFTR, indicating that its high-confidence point set exhibits significantly lower spatial dispersion and a more concentrated distribution. Although the \(ea\) is slightly larger in \#1, when considered together with its higher \(HCR\) value, it can be concluded that SGFormer matches more valid points in this scenario, albeit with a somewhat wider distribution range. SGFormer demonstrated a lower mean distance to the weighted centroid (\(md2c\)) than LoFTR across all three scenarios, with reductions of 10.0\%, 8.6\%, and 12.5\%, respectively. These consistent results indicate that the high-confidence points generated by SGFormer are more tightly clustered around the distribution centroid, exhibiting superior spatial distribution in the viewpoint change scene, which is crucial for enhancing geometric consistency in matching.

In terms of \(HCR\) and \(1Sig\) metrics, the performances of SGFormer and LoFTR varies. SGFormer shows a significant advantage in the coverage of high-confidence ratio (\(HCR\)) in \#1, demonstrating stronger feature discriminability, while LoFTR holds a slight edge in \(HCR\) for \#3 and in \(1Sig\) for \#2. It is noteworthy that SGFormer achieves a smaller \(ea\) and \(md2c\) while maintaining or improving \(1Sig\) (the coverage within one standard deviation of the ellipse), which further confirms its ability to retain a high proportion of valid points within a more compact region, reflecting a higher quality of distribution. This improvement stems from the TSA module\textquotesingle s structural guidance, which elevates distribution quality by suppressing distractions and focusing attention on geometrically coherent regions.

The consequent match distributions, visualized in Fig. 9, demonstrate that SGFormer yields a greater number of correct matches concentrated within the overlapping scene regions, whereas LoFTR produces significantly more outliers in non-overlapping areas, as is particularly evident in \#2 and \#3. Furthermore, SGFormer generates a denser set of matches compared to LoFTR, especially for image pairs exhibiting substantial illuminating and scale variation (\#1). The enhanced attentional focus above directly contributes to the improved visual localization accuracy demonstrated in Table~\ref{tab:table4}.

\subsubsection{Relative Pose Estimation on MegaDepth}

To evaluate geometric accuracy in large-scale, unstructured environments, we conducted relative pose estimation experiments on the challenging MegaDepth dataset (Li and Snavely, 2018). This dataset consists of 196 scenes reconstructed from internet photos, providing COLMAP-generated camera poses (Schonberger and Frahm, 2016). Following LoFTR (Sun et al., 2021), we used the standard test set of 1,500 image pairs from ``Sacre Coeur'' and ``St. Peter's Square'' scenes. Input images were resized to a maximum side length of 832 pixels and padded to a fixed size of 832$\times$832.

For a fair comparison, we retrained the baseline LoFTR (Sun et al., 2021) model using our identical training data (38,000 raw MegaDepth images, excluding Scene 148) and the original training hyperparameters. We compared SGFormer against this retrained LoFTR, as well as DRC-Net (Liu and Zhang, 2022) and the detector-based SuperPoint (DeTone et al., 2018) + SuperGlue (Sarlin et al., 2020). For all methods, the essential matrices were measured by the maximum angular error in rotation and translation (max($\Delta$\emph{\textbf{R}}, $\Delta$\emph{\textbf{t}})), with performance summarized as the Area Under the Cumulative error Curve (AUC) at \(5^{{^\circ}}\), \(10^{{^\circ}}\) and \(20^{{^\circ}}\) thresholds (H. Chen et al., 2022; Sun et al., 2021). Additionally, we report the Mean Reprojection Error (MRE) of the epipolar geometry and its variance across test pairs as indicators of matching consistency.

\begin{table*}[!t]
\caption{Two-view relative pose estimation results on MegaDepth-1500 test set.}
\label{tab:table6}
\centering
\input{tables/table06_v2.tex}
\vspace{1pt}
\parbox{\linewidth}{\footnotesize\emph{Note:} ``MRE'' and ``var'' represent Mean Reprojection Error of epipolar geometry and Variance of MRE, respectively.}
\end{table*}

As shown in Table~\ref{tab:table6}, SGFormer achieves superior results across all evaluation metrics. It consistently outperforms the retrained LoFTR baseline in pose estimation AUC, with improvements of 2.55\%, 2.07\%, and 0.87\% at the \(5^{{^\circ}}\), \(10^{{^\circ}}\), and \(20^{{^\circ}}\) thresholds, respectively. More importantly, SGFormer attains a significantly higher matching accuracy (98.2\%) alongside a lower and more stable reprojection (lower MRE and variance). This quantifies that SGFormer not only produces more correct matches but also yields geometrically more consistent correspondences. This superior robustness directly stems from the design of the TSA module, which reinforces feature distinctiveness within overlapping regions while suppressing distracting signals from irrelevant areas.

\begin{figure*}[!t]
\centering
\includegraphics[width=0.96\textwidth,trim={92pt 31pt 91pt 22pt},clip]{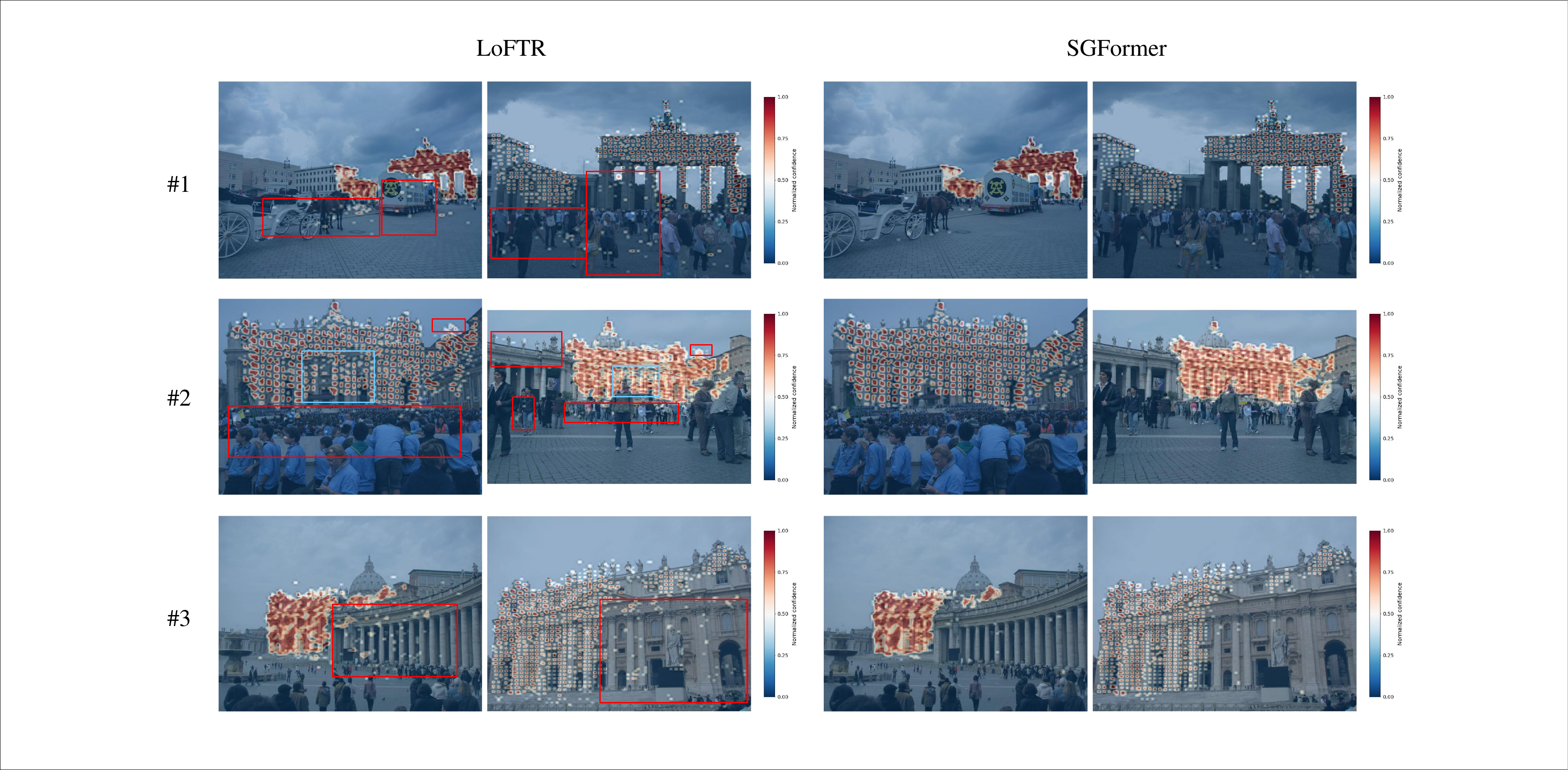}
\caption{Correlation heatmaps of LoFTR (left) and SGFormer (right) in MegaDepth dataset. Warmer color indicates higher matching confidence. The red rectangular boxes show the divergence phenomenon, and the blue rectangular boxes show the inadequate attention.}
\label{fig:fig10}
\end{figure*}

The qualitative advantages are visualized in Figs. 10 and 11. The correlation heatmaps in Fig. 10 reveal the core issue SGFormer addresses. LoFTR exhibits clear attention divergence (red boxes), scattering high-confidence responses into occluded or non-overlapping regions (e.g., sky, distant buildings). Simultaneously, it sometimes fails to concentrate sufficient attention on salient structures within truly overlapping areas (blue boxes). In contrast, SGFormer\textquotesingle s heatmaps are densely concentrated and uniformly distributed within the overlapping regions, with minimal spurious high-confidence responses elsewhere. This focused attention is a direct consequence of the structural guidance provided by the TSA module.

The quantitative results are listed in Table~\ref{tab:table7}, which corresponds to the scenes in Fig. 10. As shown in the table, SGFormer achieved a smaller fitted ellipse area (\(ea\)) in all three scenes. Compared to LoFTR, its \(ea\) values decreased by 15.0\%, 11.0\%, and 21.2\%, respectively. This result strongly demonstrates that the high-confidence point set generated by SGFormer maintains a highly concentrated spatial distribution even in complex real-world scenarios. It exhibits a stronger capability to handle large viewpoint changes and complex geometric structures, with significantly lower dispersion. Regarding the key metric for measuring the cohesion of the matched point set, \(md2c\), SGFormer also outperformed LoFTR in all scenes (with reductions of 6.5\%, 4.2\%, and 18.0\%, respectively). Particularly in \#3, the md2c reduction exceeded 18\%, indicating that SGFormer\textquotesingle s matching points are more tightly clustered spatially around the centroid. This strong aggregation directly translates to higher geometric consistency and spatial reliability of the matched point set in a viewpoint change scene.

In terms of the \(HCR\) and \(1Sig\) indicators, the performance of the two methods is comparable, with each having slight advantages in different areas. LoFTR holds a marginal advantage in \(HCR\) for \#1 and \#3, while SGFormer performs better in \(HCR\) for \#2 and in \(1Sig\) for \#1 and \#3. It is worth noting that in \#1 and \#3, SGFormer achieved a smaller distribution area (\(ea\)), a higher proportion of inlier points within the ellipse (\(1Sig\)), and a lower mean distance to the fitted ellipse centroid (\(md2c\)) with fewer confidence points (lower \(HCR\)). This further confirms the high-precision characteristic of the SGFormer: its high-confidence points are not only more reliable in distribution but also exhibit stronger spatial consistency among themselves, significantly mitigating attention divergence and demonstrating the effectiveness of the structure-guided matching design.

\begin{figure*}[!t]
\centering
\includegraphics[width=0.96\textwidth,trim={80pt 505pt 65pt 50pt},clip]{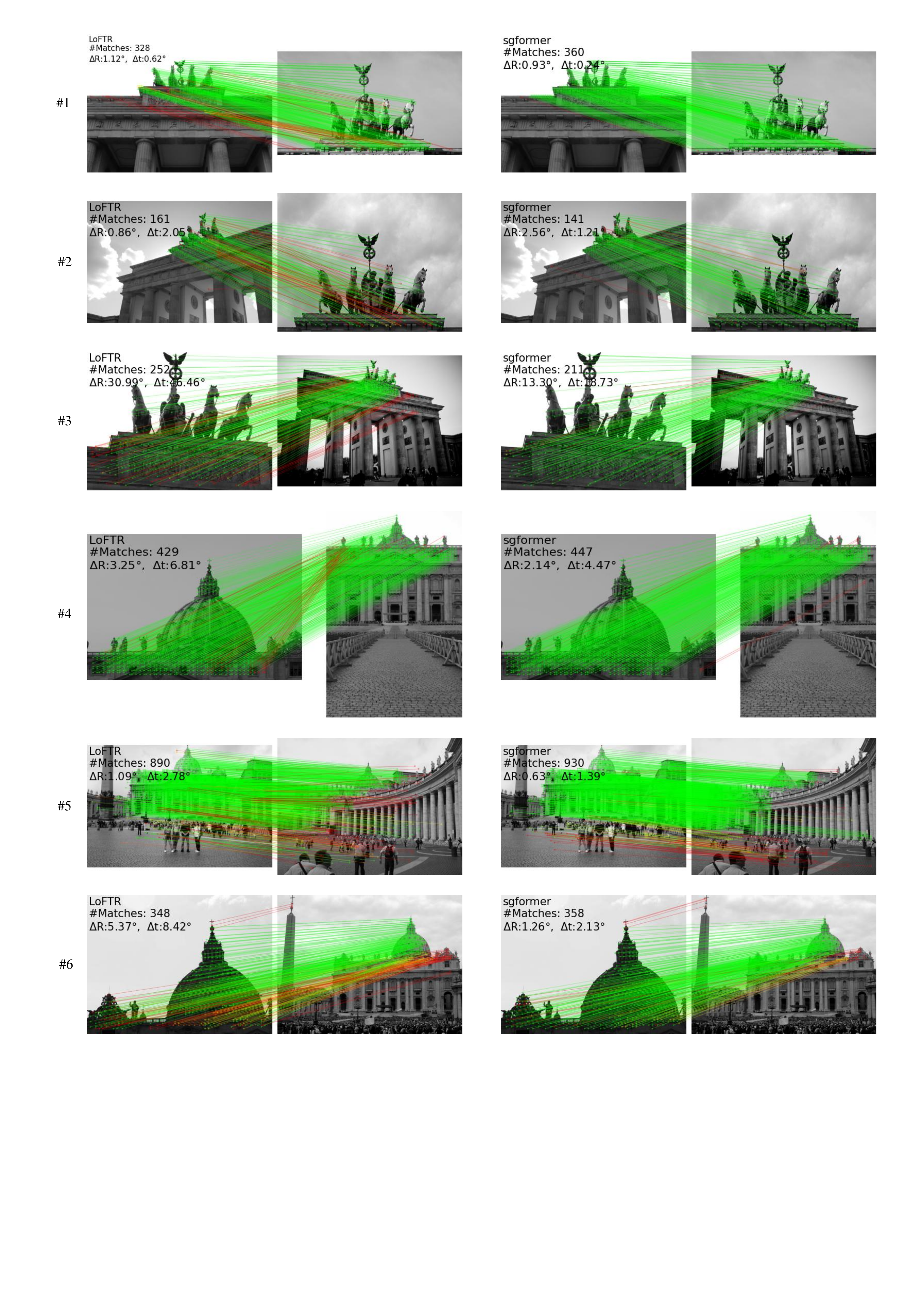}
\caption{Matching instance images of SGFormer and LoFTR on MegaDepth dataset.}
\label{fig:fig11}
\end{figure*}

\begin{table}[!t]
\caption{The quantitative results of scenes correspond to Fig. 10}
\label{tab:table7}
\centering
\input{tables/table07.tex}
\end{table}

The resulting matches, shown in Fig. 11, demonstrate SGFormer\textquotesingle s practical benefits. It generates more reliable correspondences with significantly fewer outliers across diverse challenges, including scenes with repetitive textures (\#4 \& \#6) and large-scale changes (\#6). Besides, SGFormer achieves enhanced precision in viewpoint variations (\#1: $\Delta$R/$\Delta$t = 0.93$^\circ$/0.24$^\circ$ vs. LoFTR 1.12$^\circ$/0.62$^\circ$; \#3: $\Delta$R/$\Delta$t =13.30$^\circ$/18.73$^\circ$ vs. LoFTR 30.99$^\circ$/46.46$^\circ$; \#5: $\Delta$R/$\Delta$t = 0.63$^\circ$/1.39$^\circ$ vs. LoFTR 1.09$^\circ$/2.78$^\circ$). Even in cases with slightly fewer correspondences (\#2), the higher quality and consistency of SGFormer\textquotesingle s matches ensure balanced and robust performance. These results collectively validate SGFormer\textquotesingle s efficacy in producing precise and reliable matches for accurate relative pose estimation in complex outdoor environments.

\subsection{Ablation Study}

To evaluate the contribution of key components in the proposed SGFormer architecture, we conducted an ablation study on a subset of the MegaDepth dataset comprising 1,000 randomly sampled image pairs. All variants were trained and evaluated under identical settings (data split, protocol) as described in Section 4.2.3.

The results, summarized in Table~\ref{tab:table6}, quantitatively demonstrate the importance of each designed module. First, replacing the proposed TSA module with the standard linear attention module from LoFTR leads to a significant performance drop across all pose estimation thresholds (e.g., -3.32\% AUC@5$^\circ$). This confirms that the explicit structural feature extraction and guidance provided by TSA are crucial for focusing attention and improving matching robustness. Second, removing the RoPE from the TSA module also results in noticeable accuracy degradation (-2.30\% AUC@5$^\circ$). This verifies the role of RoPE in effectively encoding rotation-equivariant spatial relationships, which aids the network in disambiguating features based on geometric context during the transformation stage. The complete SGFormer model (FULL) achieves the best performance across all metrics, indicating a synergistic effect between the structure-guided attention mechanism and the integrated geometric prior. This ablation study confirms that both the TSA module\textquotesingle s structural guidance and the RoPE\textquotesingle s spatial encoding are essential components contributing to SGFormer\textquotesingle s superior matching and pose estimation capabilities

\begin{table*}[!t]
\caption{Ablation study on the MegaDepth subset. results, Performance is measured by pose estimation AUC and matching accuracy (Acc).}
\label{tab:table8}
\centering
\input{tables/table08.tex}
\end{table*}

\section{Discussion}

This work establishes that integrating explicit structural similarity guidance into a detector-free matching framework significantly enhances feature representation within overlapping regions. Experimental validation confirms that SGFormer achieves excellent performance, particularly under challenging photometric variations, where it attains 11.76\% improvement in illumination-invariant matching accuracy at 1px threshold compared to LoFTR (Table~\ref{tab:table2}). This advancement, alongside the best overall matching precision, is substantiated by the visual evidence in Figs. 7-11. Experimental results show that shallow structural features effectively concentrate the network\textquotesingle s attention on geometrically consistent areas, thereby attenuating interference from photometric distractors and mitigating the attention divergence prevalent in unconstrained Transformer models.

Compared to LoFTR, our approach more effectively focuses attention on valid matching regions while suppressing interference from irrelevant features, leading to enhanced matching precision and robustness. Furthermore, in contrast to methods that incorporate dedicated preprocessing modules for overlap estimation (e.g., OETR), SGFormer employs a lightweight, integrated strategy. It utilizes light-weight layers to extract shallow structural features for guidance directly within the feature transformation process. This design avoids the substantial increases in model parameters and training complexity associated with auxiliary functional networks.

It should be noted that while SGFormer achieves higher overall matching accuracy and superior pose estimation on MegaDepth compared to LoFTR, performance can vary in specific scenarios. For instance, in isolated cases like that illustrated in Fig. 11 (\#2), LoFTR may yield a marginally lower pose error despite SGFormer\textquotesingle s higher match accuracy. Analysis suggests this occurs when LoFTR, by chance, produces a smaller set of matches that are precisely localized at highly distinctive, salient points (e.g., corners), which are optimal for pose estimation. This observation highlights a nuanced interplay between match quantity and geometric salience. Additionally, we observe that the effectiveness of structural similarity guidance can be compromised under extreme viewpoint changes involving significant scale variations or non-rigid deformations, as the geometric consistency of the guiding features themselves becomes less reliable. This represents a primary limitation of the current approach.

Future work will focus on enhancing the geometric invariance of the guidance mechanism. One direction involves designing lightweight modules to extract multi-scale shallow structural features that are more robust to scale changes and perspective distortions, aiming to preserve guidance effectiveness and increase reliable matches in extreme viewing conditions. Another direction targets scenes with highly repetitive textures. Here, constructing an efficient semantic relation model that strengthens both relative and absolute spatial-semantic awareness could help disambiguate visually similar features by leveraging higher-level contextual information. These advancements seek to further improve the robustness and applicability of structure-guided matching in the most demanding photogrammetric scenarios.

\section{Conclusion}

In this paper, we propose SGFormer, a novel semi-dense local feature matching framework designed to address the critical problem of attention divergence in challenging photogrammetric scenarios. The core of our method is a hierarchical Transformer backbone guided by explicit structural similarity, which systematically directs the network's focus toward overlapping regions. The key innovation is the Triple-Structure-Attention (TSA) module, which utilizes high-resolution shallow structural features to guide the aggregation of discriminative visual representations in early network stages. These structurally-primed features are then progressively refined in deeper layers through interleaved self- and cross-attention, ensuring that the model maintains a concentrated focus on corresponding areas while suppressing interference from irrelevant regions. Integrated into a semi-dense coarse-to-fine matching pipeline, this design enables robust correspondence estimation with sub-pixel accuracy. Comprehensive experiments demonstrate that SGFormer effectively mitigates the attention divergence prevalent in existing detector-free methods. It achieves excellent performance in image matching--for instance, leading the second place by 4.84\%, 1.10\%, and 2.15\% at the 1px, 3px, and 5px thresholds on HPatches, respectively, and attaining the highest matching accuracy (98.2\%) on MegaDepth-1500. Furthermore, it exhibits superior robustness in downstream photogrammetric tasks such as visual localization and relative pose estimation. For example, it achieves the best localization performance under the (0.25m, 2$^\circ$) threshold in night conditions on Aachen-Day-Night, and surpasses the second-best method by 2.55\%, 2.07\%, and 0.87\% at the \(5^{{^\circ}}\), \(10^{{^\circ}}\), and \(20^{{^\circ}}\) thresholds on MegaDepth. These results validate that explicitly incorporating structural guidance into the matching process significantly enhances both the precision and reliability of image correspondences, advancing the capability for high-fidelity 3D reconstruction, stereo mapping, and bundle adjustment.

\appendices
\section*{Appendix}

\begin{figure*}[!t]
\centering
\includegraphics[width=0.96\textwidth,trim={103pt 15pt 98pt 27pt},clip]{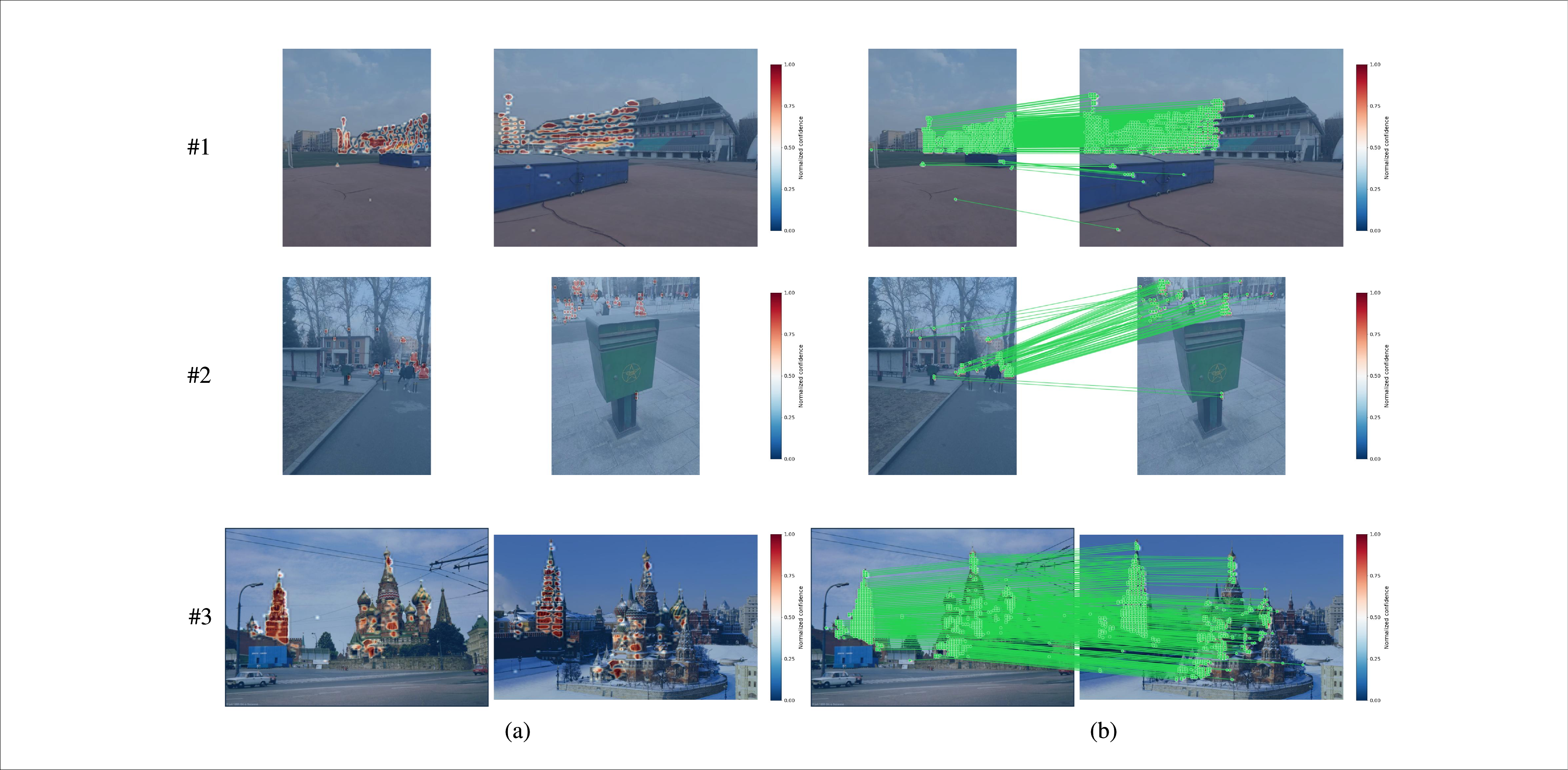}
\caption{The correlation heatmap achieved by the proposed SGFormer (Column a), and its corresponding putative matches (Column b). Compared with the visualization in Fig. 1 at Sec. Introduction, SGFormer shows concentrated attention with better matching accuracy and reliability.}
\label{fig:fig12}
\end{figure*}

Additional visualization results in Fig.~\ref{fig:fig12} further demonstrate that SGFormer produces more concentrated attention responses and more reliable putative matches than LoFTR.

\end{document}

%% file: tables/table01.tex
\begingroup
\footnotesize
\renewcommand{\arraystretch}{1.15}
\setlength{\tabcolsep}{4pt}
\newcommand{\datasetexample}[1]{%
  \IfFileExists{#1}{%
    \includegraphics[width=0.96\linewidth,height=0.72in,keepaspectratio]{#1}%
  }{%
    \fbox{\parbox[c][0.58in][c]{0.92\linewidth}{\centering Example image placeholder\\[-1pt]\texttt{\detokenize{#1}}}}%
  }%
}
\begin{tabular}{@{}>{\raggedright\arraybackslash}p{0.13\linewidth} >{\raggedright\arraybackslash}p{0.83\linewidth}@{}}
\toprule
\textbf{Settings} & \textbf{Information} \\
\midrule
\multirow{12}{*}{\textbf{Datasets}}
& (1) HPatches dataset, 108 scene sequences (52 illumination, 56 viewpoint change) for eval image matching.\\
& \url{https://github.com/hpatches/hpatches-dataset}\\[-1pt]
& \textit{Example images:}\\[-1pt]
& \datasetexample{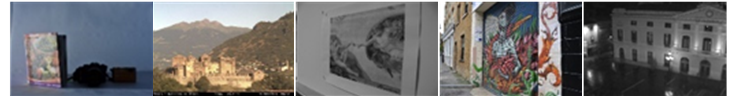}\\[4pt]
\cmidrule(l){2-2}
& (2) MegaDepth dataset, 1500 image pairs in ``Sacre Coeur'' and ``St. Peter's Square'' for eval relative pose estimating.\\
& \url{https://www.cs.cornell.edu/projects/megadepth/}\\[-1pt]
& \textit{Example images:}\\[-1pt]
& \datasetexample{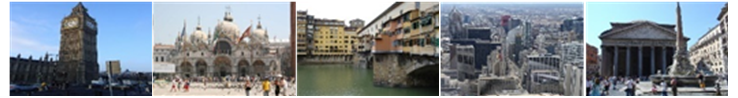}\\[4pt]
\cmidrule(l){2-2}
& (3) Aachen-Day-Night dataset, 4328 reference images and 922 query images for eval visual localization.\\
& \url{https://data.ciirc.cvut.cz/public/projects/2020VisualLocalization/Aachen-Day-Night/}\\[-1pt]
& \textit{Example images:}\\[-1pt]
& \datasetexample{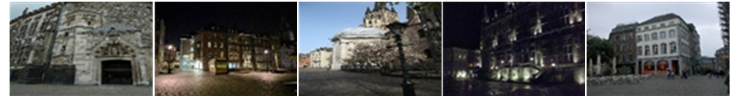}\\
\bottomrule
\end{tabular}
\endgroup

%% file: tables/table02.tex
\begingroup
\footnotesize
\renewcommand{\arraystretch}{1.08}
\setlength{\tabcolsep}{2.5pt}
\begin{tabular}{@{}ll@{\hspace{4pt}}ccc@{}}
\toprule
\textbf{Category} & \textbf{Method} & \multicolumn{3}{c}{\textbf{Mean Matching Accuracy (MMA)}} \\
\cmidrule(lr){3-5}
& & \textbf{Overall} & \textbf{Illumination} & \textbf{Viewpoint} \\
\midrule
\multirow{4}{*}{Detector-based}
& CAPS\_SIFT+NN & 0.33/0.64/0.74 & 0.42/0.68/0.78 & 0.24/0.61/0.70 \\
& SuperPoint+SuperGlue & 0.44/0.85/0.95 & 0.50/0.86/0.96 & 0.39/0.85/\textbf{0.93} \\
& ASLFeat+NN & 0.40/0.72/0.82 & 0.47/0.77/0.88 & 0.33/0.67/0.77 \\
& D2-Net & 0.11/0.43/0.67 & 0.17/0.53/0.77 & 0.06/0.34/0.57 \\
\midrule
\multirow{5}{*}{Detector-free}
& COTR & 0.38/0.68/0.77 & 0.39/0.65/0.75 & 0.37/0.70/0.79 \\
& Sparse-NCNet & 0.45/0.79/0.87 & 0.62/0.84/0.91 & 0.29/0.73/0.83 \\
& Patch2Pix & 0.41/0.81/0.90 & 0.53/0.90/0.97 & 0.29/0.73/0.84 \\
& LoFTR & 0.62/0.91/0.93 & 0.68/0.96/\textbf{0.98} & \textbf{0.57}/0.85/0.89 \\
& SGFormer (ours) & \textbf{0.65/0.92/0.95} & \textbf{0.76/0.98/0.98} & 0.55/\textbf{0.86}/0.91 \\
\bottomrule
\end{tabular}
\endgroup

%% file: tables/table03.tex
\begingroup
\footnotesize
\renewcommand{\arraystretch}{1.12}
\setlength{\tabcolsep}{6pt}
\begin{tabular}{@{}ccc@{}}
\toprule
\textbf{\#} & \textbf{LoFTR HCR$\uparrow$} & \textbf{SGFormer HCR$\uparrow$} \\
\midrule
1 & 0.204 & \textbf{0.554} \\
2 & 0.593 & \textbf{0.645} \\
3 & 0.524 & \textbf{0.541} \\
\bottomrule
\end{tabular}
\endgroup

%% file: tables/table04.tex
\begingroup
\footnotesize
\renewcommand{\arraystretch}{1.2}
\setlength{\tabcolsep}{4pt}
\begin{tabular*}{0.88\textwidth}{@{\extracolsep{\fill}}llcc@{}}
\toprule
\textbf{Category} & \textbf{Method} & \textbf{Day} & \textbf{Night} \\
\midrule
\multirow{2}{*}{Detector-based}
& D2-Net & -- & 74.5/86.7/\textbf{100.0} \\
& SuperPoint+SuperGlue & \textbf{89.3/95.9/98.8} & \textbf{87.8}/\underline{92.9}/\textbf{100.0} \\
\midrule
\multirow{5}{*}{Detector-free}
& ASpanFormer & 88.7/94.1/98.1 & \textbf{87.8/93.9/100.0} \\
& COTR & 82.4/91.9/96.8 & 75.5/90.8/\underline{99.0} \\
& LoFTR & \underline{89.2}/94.5/\underline{98.3} & \underline{86.7}/91.8/\underline{99.0} \\
& TopicFM & 88.8/94.7/97.9 & \underline{86.7}/\underline{92.9}/\textbf{100.0} \\
& SGFormer (ours) & 88.7/\underline{95.0}/97.9 & \textbf{87.8}/\underline{92.9}/98.0 \\
\bottomrule
\end{tabular*}
\endgroup

%% file: tables/table05.tex
\begingroup
\scriptsize
\renewcommand{\arraystretch}{1.12}
\setlength{\tabcolsep}{1pt}
\begin{tabular}{@{}c@{\hspace{2pt}}>{\centering\arraybackslash}p{0.088\columnwidth}>{\centering\arraybackslash}p{0.118\columnwidth}>{\centering\arraybackslash}p{0.088\columnwidth}>{\centering\arraybackslash}p{0.102\columnwidth}@{\hspace{5pt}}>{\centering\arraybackslash}p{0.088\columnwidth}>{\centering\arraybackslash}p{0.118\columnwidth}>{\centering\arraybackslash}p{0.088\columnwidth}>{\centering\arraybackslash}p{0.102\columnwidth}@{}}
\toprule
\multirow{2}{*}{\textbf{\#}} & \multicolumn{4}{c}{\textbf{LoFTR}} & \multicolumn{4}{c}{\textbf{SGFormer}} \\
\cmidrule(lr){2-5}\cmidrule(lr){6-9}
& \textbf{HCR$\uparrow$} & \textbf{ea$\downarrow$} & \textbf{1Sig$\uparrow$} & \textbf{md2c$\downarrow$} & \textbf{HCR$\uparrow$} & \textbf{ea$\downarrow$} & \textbf{1Sig$\uparrow$} & \textbf{md2c$\downarrow$} \\
\midrule
1 & 0.072 & \textbf{179574} & 0.303 & 334.1 & \textbf{0.256} & 180466 & \textbf{0.347} & \textbf{300.6} \\
2 & 0.117 & 114796 & \textbf{0.345} & 254.8 & \textbf{0.151} & \textbf{99428} & 0.337 & \textbf{232.9} \\
3 & \textbf{0.121} & 137597 & 0.423 & 286.7 & 0.119 & \textbf{114502} & \textbf{0.440} & \textbf{250.9} \\
\bottomrule
\end{tabular}
\endgroup

%% file: tables/table06_v2.tex
\begingroup
\footnotesize
\renewcommand{\arraystretch}{1.10}
\setlength{\tabcolsep}{2pt}
\begin{tabular}{@{}>{\raggedright\arraybackslash}p{0.14\textwidth}>{\raggedright\arraybackslash}p{0.23\textwidth}*{3}{>{\centering\arraybackslash}p{0.075\textwidth}}>{\centering\arraybackslash}p{0.06\textwidth}>{\centering\arraybackslash}p{0.13\textwidth}>{\centering\arraybackslash}p{0.13\textwidth}@{}}
\toprule
\textbf{Category} & \textbf{Method} & \multicolumn{3}{c}{\textbf{Relative Pose Est. AUC}} & \textbf{Acc} & \textbf{MRE$\downarrow$} & \textbf{Var$\downarrow$} \\
\cmidrule(lr){3-5}
& & \textbf{$5^{\circ}$} & \textbf{$10^{\circ}$} & \textbf{$20^{\circ}$} & & & \\
\midrule
Detector-based
& SuperPoint+SuperGlue & 42.2 & 61.2 & 76.1 & -- & -- & -- \\
\midrule
\multirow{3}{=}{Detector-free}
& DRC-Net & 27.1 & 43.2 & 58.4 & -- & -- & -- \\
& LoFTR & 50.9 & 67.5 & 80.3 & 0.962 & $2.012\times10^{-4}$ & $3.518\times10^{-5}$ \\
& SGFormer (ours) & \textbf{52.2} & \textbf{68.9} & \textbf{81.0} & \textbf{0.982} & $\mathbf{1.433\times10^{-4}}$ & $\mathbf{2.278\times10^{-5}}$ \\
\bottomrule
\end{tabular}
\endgroup

%% file: tables/table07.tex
\begingroup
\scriptsize
\renewcommand{\arraystretch}{1.12}
\setlength{\tabcolsep}{1pt}
\begin{tabular}{@{}c@{\hspace{2pt}}>{\centering\arraybackslash}p{0.088\columnwidth}>{\centering\arraybackslash}p{0.118\columnwidth}>{\centering\arraybackslash}p{0.088\columnwidth}>{\centering\arraybackslash}p{0.102\columnwidth}@{\hspace{5pt}}>{\centering\arraybackslash}p{0.088\columnwidth}>{\centering\arraybackslash}p{0.118\columnwidth}>{\centering\arraybackslash}p{0.088\columnwidth}>{\centering\arraybackslash}p{0.102\columnwidth}@{}}
\toprule
\multirow{2}{*}{\textbf{\#}} & \multicolumn{4}{c}{\textbf{LoFTR}} & \multicolumn{4}{c}{\textbf{SGFormer}} \\
\cmidrule(lr){2-5}\cmidrule(lr){6-9}
& \textbf{HCR$\uparrow$} & \textbf{ea$\downarrow$} & \textbf{1Sig$\uparrow$} & \textbf{md2c$\downarrow$} & \textbf{HCR$\uparrow$} & \textbf{ea$\downarrow$} & \textbf{1Sig$\uparrow$} & \textbf{md2c$\downarrow$} \\
\midrule
1 & \textbf{0.158} & 117675 & 0.389 & 314.6 & 0.137 & \textbf{100065} & \textbf{0.414} & \textbf{294.1} \\
2 & 0.252 & 93900 & \textbf{0.302} & 270.7 & \textbf{0.258} & \textbf{83619} & 0.284 & \textbf{259.3} \\
3 & \textbf{0.211} & 124577 & 0.377 & 293.1 & 0.177 & \textbf{98233} & \textbf{0.393} & \textbf{240.3} \\
\bottomrule
\end{tabular}
\endgroup

%% file: tables/table08.tex
\begingroup
\footnotesize
\renewcommand{\arraystretch}{1.15}
\setlength{\tabcolsep}{3.5pt}
\begin{tabular}{@{}>{\raggedright\arraybackslash}p{0.47\linewidth}
                >{\centering\arraybackslash}p{0.105\linewidth}
                >{\centering\arraybackslash}p{0.105\linewidth}
                >{\centering\arraybackslash}p{0.105\linewidth}
                >{\centering\arraybackslash}p{0.09\linewidth}@{}}
\toprule
\textbf{Method} & \multicolumn{3}{c}{\textbf{Pose Estimation AUC}} & \textbf{Acc} \\
\cmidrule(lr){2-4}
& \textbf{$5^{\circ}$} & \textbf{$10^{\circ}$} & \textbf{$20^{\circ}$} & \\
\midrule
1) replace TSAM with Linear Attention module & 0.379 & 0.539 & 0.667 & 0.851 \\
2) remove exclusively embedded RoPE & 0.383 & 0.537 & 0.665 & 0.854 \\
3) FULL & \textbf{0.392} & \textbf{0.555} & \textbf{0.679} & \textbf{0.861} \\
\bottomrule
\end{tabular}
\endgroup